\RequirePackage{etex}
\documentclass{aa}

\usepackage{graphicx}
\usepackage{txfonts}
\usepackage{amsmath}
\usepackage{booktabs}
\usepackage{placeins}
\usepackage{needspace}
\usepackage[table]{xcolor}
\definecolor{pcgsrow}{HTML}{FBECE0}
\usepackage[colorlinks=true,citecolor=blue,linkcolor=blue,urlcolor=blue]{hyperref}
\expandafter\def\expandafter\UrlBreaks\expandafter{\UrlBreaks\do\/\do-\do\_}
\renewcommand{\topfraction}{0.92}

\renewcommand{\floatpagefraction}{0.88}

\graphicspath{{figs/}{../results/figures/nc/}}
\newif\iflightfigs \lightfigsfalse  
\newcommand{\figfile}[1]{\iflightfigs figs_light/#1.jpg\else #1.pdf\fi}

\newcommand{\pcgs}{PCGS}
\newcommand{\mzero}{\mu_0}
\newcommand{\vect}[1]{\boldsymbol{#1}}
\makeatletter
\newcommand{\settablecaptype}{\def\@captype{table}}
\newcommand{\setfigcaptype}{\def\@captype{figure}}
\makeatother

\begin{document}

\title{PCGS: Photometrically constrained Gaussian splatting for sparse-view asteroid shape reconstruction}

\titlerunning{PCGS: Gaussian splatting for sparse-view asteroid shape reconstruction}
\authorrunning{X.-Y. Xiao et al.}

\author{Xin-Yu Xiao\inst{1,2}
        \and Fei He\inst{1,2}\thanks{Corresponding author: \email{hefei@mail.iggcas.ac.cn}}
        \and Wei Lin\inst{1,2}}

\institute{Institute of Geology and Geophysics, Chinese Academy of Sciences, Beijing 100029, China
           \and
           College of Earth and Planetary Sciences, University of Chinese Academy of Sciences, Beijing 100049, China}

\date{Received ...; accepted ...}

\abstract
{Global asteroid shape models are needed within days of spacecraft arrival from only a few dozen images. Gaussian
splatting builds surfaces in tens of minutes, but models brightness with view-dependent colours that absorb topographic
shading rather than using the known solar illumination.}
{We use the known solar illumination directly to constrain surface orientation in Gaussian splatting, preserve unlit
terrain across wide phase angles, and validate the closed models against mission references, including laser altimetry.}
{In photometrically constrained Gaussian splatting (\pcgs), planar surfels carry a view-independent scalar albedo
shaded by a lunar-Lambert law under each image's solar direction. A scaled night-side band excludes the unlit limb from
the silhouette loss. A phase switch selects an albedo total-variation prior ($\alpha<15^\circ$) or detached Sun-view
cast shadows ($\alpha\geq 15^\circ$) before depth-consensus meshing. We benchmarked \pcgs\ against six baselines (2DGS,
GOF, RaDe-GS, PGSR, Asteroid-GS, and COLMAP dense MVS) across five datasets of Itokawa, Ryugu, and Bennu
($\alpha=7^\circ$--$60^\circ$).}
{On Itokawa, \pcgs\ attains 0.59~m mean and Chamfer distance, outperforming all baselines including RaDe-GS (0.62~m)
and COLMAP dense MVS (0.85~m). On Ryugu ($19^\circ$), \pcgs\ reaches 0.62~m, 15\% below PGSR, 45\% below COLMAP dense
MVS, and five times below RaDe-GS (3.12~m). Against Bennu's laser-altimeter model, \pcgs\ reaches 0.50~m,
matching COLMAP dense MVS (0.53~m) in mean error while better preserving boulder-scale surface area and synthesising
novel views, whereas all competing Gaussian-splatting baselines exceed 1.2~m mean and 3.7~m Chamfer. Reverting to
view-dependent colour raises first-split mean error by 30\% on Itokawa, 6\% on Ryugu ($19^\circ$), and from 0.51 to
4.7~m on Bennu; omitting the night-side band raises Bennu error to 4.2~m. Enclosed volumes agree within 0.38\%
after registration and 3.4\% at nominal trajectory scale (1.1\% on Ryugu and Bennu).}
{Combining lunar-Lambert shading, terminator protection, and depth fusion yields watertight sub-metre shape models from
roughly 50 early-rendezvous images in about 30~min on one GPU.}


\maketitle
\nolinenumbers

\section{Introduction}
\label{sec:intro}

A global three-dimensional shape model is among the first products needed when a spacecraft arrives at a small body.
Combined with radio-science mass estimates, its enclosed volume gives the bulk density and macroporosity that
identified (25143) Itokawa \citep{Fujiwara2006} and (162173) Ryugu \citep{Watanabe2019} as rubble piles. Global
topography also supports gravity-field, slope, and thermal modelling, optical navigation, and sampling-site hazard
assessment \citep{Lauretta2019}. During early arrival on missions such as ESA's Hera \citep{Michel2022}, teams
therefore need a closed global model rapidly from only a few dozen images.

Before arrival, ground-based lightcurve inversion \citep{Kaasalainen2001} and delay-Doppler radar \citep{Ostro2002}
constrain spin states and coarse shapes, but lack the resolution required for proximity operations. Once in situ
images are acquired, mission models are traditionally built by stereo\-photo\-gram\-metry (SPG;
\citealt{Preusker2015,Watanabe2019}) or stereo\-photo\-clino\-metry (SPC; \citealt{Gaskell2008,Palmer2022}). SPG
triangulates features across images with similar lighting, whereas SPC combines stereo control with
multi-illumination photoclinometry to infer surface slopes from brightness. Both achieve high accuracy and remain the
standard for late-mission global products, but require hundreds to thousands of images -- and, for SPC, months of
analyst effort -- precluding rapid characterisation during the initial days after arrival.

Recent sparse-view approaches therefore turn to neural rendering. Neural implicit surface methods reach metre-scale
accuracy from tens of images on Itokawa, Bennu, and Ryugu \citep{ChenH2024,ChenS2024,ChenH2025}, but require hours
to days of training. Three-dimensional Gaussian splatting \citep{Kerbl2023} and surface-oriented variants
\citep{Huang2024,Yu2024GOF,Chen2024PGSR} cut training to tens of minutes, and Asteroid-GS
\citep{Zhang2025AsteroidGS} adapted this framework to asteroids. Unlike SPC, however, none of these methods enforces
a physical reflectance law: general pipelines use unconstrained view-dependent spherical harmonics, while
Asteroid-GS uses illumination-conditioned colour networks and per-image codes. On airless regolith, such flexible
appearance models can absorb topographic shading into colour rather than constraining surface normals.

Using planetary photometry as a global constraint in Gaussian splatting introduces three phase-dependent challenges.
First, near opposition (solar phase angle $\alpha\lesssim 10^\circ$), weak shading contrast makes local albedo and
surface slope difficult to separate. Second, at moderate to high phase angles ($\alpha\gtrsim 15^\circ$), boulders
and craters cast shadows that a local reflectance law cannot reproduce. Third, at high phase angles
($\alpha\gtrsim 30^\circ$), unlit terrain beyond the terminator is missing from brightness-based foreground masks
and is carved away by a standard silhouette loss.

Here we present photometrically constrained Gaussian splatting (\pcgs), which integrates planetary reflectance into
two-dimensional Gaussian splatting (2DGS; \citealt{Huang2024}) to complement late-mission SPG, SPC, and altimetry
during the early-arrival window. Each planar surfel carries a view-independent relative scalar albedo shaded by the
lunar-Lambert law under the known solar direction of each image, directly constraining surface normals. To address all three phase regimes, \pcgs\ excludes a scaled night-side band beyond the terminator from the
silhouette loss and switches between an albedo total-variation regulariser ($\alpha<15^\circ$) and detached
(stop-gradient) Sun-view cast shadows ($\alpha\geq 15^\circ$) before sunlit-consistent depth-consensus meshing. We
benchmark \pcgs\ against six baselines -- five surface-oriented Gaussian-splatting methods and classical COLMAP dense
multi-view stereo (MVS) -- across five datasets of Itokawa, Ryugu, and (101955) Bennu ($\alpha=7^\circ$--$60^\circ$),
including Bennu's laser-altimeter reference.

Section~\ref{sec:related} reviews related work, Sect.~\ref{sec:method} formulates \pcgs, Sect.~\ref{sec:setup}
describes the datasets and evaluation protocol, Sect.~\ref{sec:results} presents the benchmarks and ablations,
Sect.~\ref{sec:discussion} discusses accuracy and limitations, and Sect.~\ref{sec:conclusions} summarises our
conclusions.

\section{Related work}
\label{sec:related}

\subsection{Shape modelling of small bodies}

In mission practice, stereo geometry and structure-from-motion multi-view stereo (SfM-MVS) pipelines
\citep{Furukawa2010,Schonberger2016MVS} underlie SPG and SfM-MVS shape models of (4) Vesta \citep{Jaumann2012}, comet 67P/Churyumov--Gerasimenko
\citep{Preusker2015,Preusker2017}, and Ryugu \citep{Watanabe2019}. Where multi-illumination coverage is available, SPC builds on photoclinometry \citep{Horn1990}, integrating dense
networks of local digital terrain maps (maplets) to derive global models such as those of Itokawa \citep{Gaskell2008},
comet 67P \citep{Jorda2016}, (1) Ceres \citep{Park2019}, and Bennu \citep{Palmer2022}, while multi-view
shape-from-shading is similarly combined with stereo or altimetry to refine local planetary terrain models
\citep{Kirk2003,Grumpe2014,Alexandrov2018}. When a laser altimeter is carried, active ranging yields an
image-independent benchmark: the OSIRIS-REx Laser Altimeter \citep[OLA;][]{Daly2017,Barnouin2019,AlAsad2021}
produced a standalone global model of Bennu \citep{Seabrook2022}. All of these mission products, however, rely on
dense orbital coverage or dedicated ranging payloads.

\subsection{Surface reconstruction by neural rendering}

Building on neural radiance fields \citep{Mildenhall2020} and neural signed distance functions
\citep{Wang2021NeuS,Li2023Neuralangelo}, planetary adaptations have reconstructed Itokawa, Bennu, and Ryugu from
tens of images \citep{ChenH2024,ChenS2024,ChenH2025}, albeit with multi-hour optimisation or auxiliary SfM-MVS
supervision. By contrast, volumetric Gaussian splatting \citep{Kerbl2023} and anti-aliased Mip-Splatting \citep{Yu2024Mip} focus
on view synthesis rather than accurate geometry. Surface-oriented variants enforce geometric structure by binding
primitives to meshes \citep{Guedon2024}, flattening Gaussians into planar surfels \citep{Huang2024,Dai2024Surfels},
extracting surfaces from Gaussian Opacity Fields \citep[GOF;][]{Yu2024GOF} or rasterised depths
\citep{Zhang2024RaDe}, or adding multi-view patch consistency in Planar-based Gaussian Splatting Reconstruction
\citep[PGSR;][]{Chen2024PGSR}. Asteroid-GS \citep{Zhang2025AsteroidGS} introduces dilated-mask pruning,
illumination-conditioned multilayer perceptrons (MLPs), and geometric regularisers, yet its shadow term approximates
solar incidence from the body-centroid radial direction rather than the local normal. Across all of these pipelines,
appearance remains parametrised by spherical harmonics or per-frame codes \citep{MartinBrualla2021} rather than a
physical reflectance law.

\subsection{Illumination in neural scene representations}

In Earth observation, satellite neural radiance fields \citep{Derksen2021,Mari2022,Mari2023} and Earth Observation
Gaussian Splatting \citep[EOGS;][]{SavantAira2025} use known solar angles and shadow mapping, but model open local
terrain patches without a closed 3D limb. Terrestrial inverse rendering
\citep{Zhang2021NeRFactor,Liang2024GSIR,Jiang2024GaussianShader,Gao2024Relightable} instead solves jointly for
unknown environment lighting and material properties rather than exploiting a collimated solar source. Closest to
our work, \citet{Nolan2026} replaced spherical harmonics with analytical reflectance laws on local terrain tiles of
Vesta and Ceres, obtaining the best surface normals with Lommel--Seeliger shading. Although planetary reflectance
laws such as lunar-Lambert \citep{McEwen1991} and Hapke \citep{Hapke2012} are well established \citep[reviewed
by][]{Li2015}, extending them from local tiles to closed global asteroid reconstruction requires handling cast
shadows and preventing silhouette carving of the unlit night-side limb.

\section{Method}
\label{sec:method}

\begin{figure*}[!tbp]
\centering
\includegraphics[width=\textwidth]{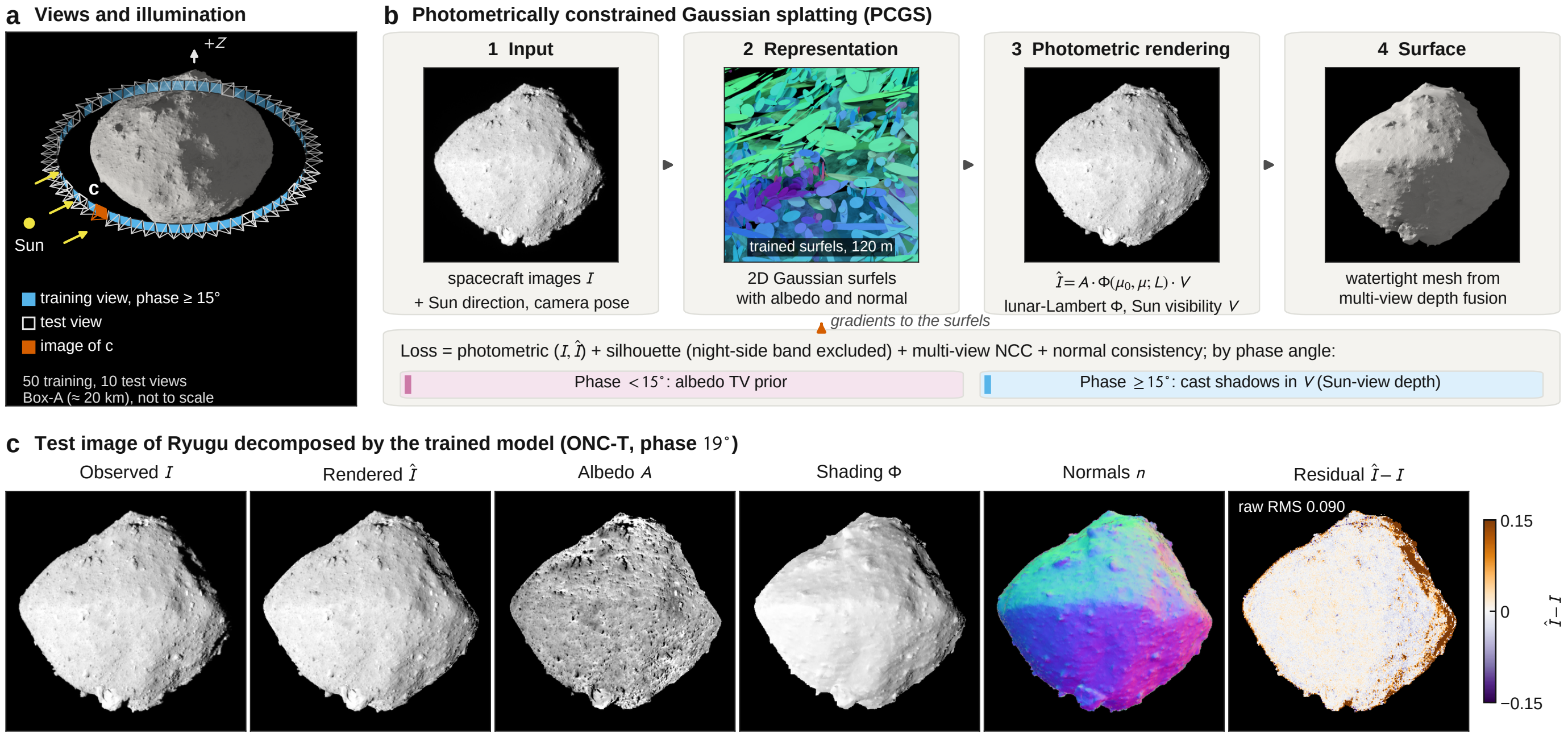}
\caption{Overview of \pcgs. (a) Ryugu reference shape model showing the $+Z$ spin axis and illuminated as in (c),
with 20~km Box-A camera poses placed along their body-fixed viewing directions (ranges compressed): training views
(blue), test views (white), test camera of (c) (orange), and Sun vectors (yellow). (b) Pipeline: images, poses, and
SPICE solar vectors supervise 2D Gaussian surfels with scalar albedos (inset: trained surfels over 120~m, coloured
by normal) rendered with Eq.~\eqref{eq:render} before depth-consensus meshing. (c) Intrinsic decomposition of a
held-out Ryugu test view (Hayabusa2 ONC-T, $\alpha=19^\circ$) into rendered brightness $\hat I$, albedo $A$,
lunar-Lambert shading $\Phi$, normals $\vect{n}$, and residual $\hat I - I$ (raw RMS $0.090$).}
\label{fig:overview}
\end{figure*}

Figure~\ref{fig:overview} gives an overview of \pcgs. Built on 2DGS \citep{Huang2024}, it keeps the differentiable
surfel rasteriser, adaptive densification, and normal--depth regularisation while replacing unconstrained
view-dependent colour with planetary photometry. The inputs are monochrome spacecraft images, body-fixed
camera poses refined by COLMAP bundle adjustment, intensity-based foreground masks covering only the sunlit disc,
and unit solar vectors from SPICE kernels \citep{Acton1996}. The surface is represented by planar surfels
(Sect.~\ref{sec:2dgs}) rendered under the solar geometry of each image (Sect.~\ref{sec:render}).
Section~\ref{sec:priors} describes the night-side band and phase switch, Sect.~\ref{sec:losses} presents the full
training loss, and Sect.~\ref{sec:extraction} explains how a watertight mesh is extracted by depth consensus.

\subsection{Two-dimensional Gaussian splatting}
\label{sec:2dgs}

In 2DGS, a scene is represented by oriented planar Gaussian surfels, each defined by a centre, two orthogonal
tangent vectors with spatial scales, a peak opacity, and an appearance attribute \citep{Huang2024}. Along each
camera ray, surfels are sorted by depth and alpha-composited with weights
\begin{equation}
w_k = o_k \prod_{j<k}\left(1-o_j\right),
\label{eq:weights}
\end{equation}
where $o_k$ is the Gaussian-weighted opacity of surfel $k$ at the ray intersection. The same weights yield the
composited surface normal $\vect{n}$ and the accumulated opacity $O=\sum_k w_k$. We take the median termination
depth where $O=0.5$, rather than the expected ray depth, because expected depth blends foreground limb surfels with
background primitives along grazing silhouettes against dark sky and biases the limb geometry.

\subsection{Photometric rendering}
\label{sec:render}

\pcgs\ replaces spherical harmonics with a single view-independent relative scalar albedo $a_k$ per surfel. Because the
input images are linearly scaled per dataset or left in archived digital numbers without absolute radiometric
calibration (Sect.~\ref{sec:images}), $a_k$ represents a relative pseudo-albedo absorbing the dataset brightness
scale and campaign-mean phase function rather than a physical normal or Bond albedo. The composited relative albedo
$A=\sum_k w_k a_k$ is shaded at each pixel as
\begin{equation}
\hat I = A\,\Phi(\mzero,\mu)\,V,
\label{eq:render}
\end{equation}
using the lunar-Lambert disc function \citep{McEwen1991}:
\begin{equation}
\Phi(\mzero,\mu) = (1-L)\,\mzero + \frac{2L\,\mzero}{\mzero+\mu}.
\label{eq:ll}
\end{equation}
Here $\vect{s}$ is the SPICE body-fixed unit solar vector \citep{Acton1996}, $\vect{v}$ is the surface-to-camera
unit ray, the incidence and emission cosines are $\mzero=\max(\vect{n}\cdot\vect{s},0)$ and
$\mu=\max(\vect{n}\cdot\vect{v},10^{-3})$ using the unit normal $\vect{n}$, and $V\in[0,1]$ is the solar visibility.
Equation~\eqref{eq:ll} interpolates between Lambertian ($L=0$) and Lommel--Seeliger ($L=1$) scattering ($\Phi=1$ at
$\mzero=\mu=1$), with $L\in(0,1)$ learned jointly as a single campaign-wide scalar. Equation~\eqref{eq:ll} includes no
explicit solar phase function $f(\alpha)$; on narrow-phase campaigns this factor is absorbed into $A$, whereas on
wide-phase sequences such as Bennu ($\alpha=26^\circ$--$60^\circ$), $L$ acts as an effective parameter that also
absorbs inter-frame phase darkening (Sect.~\ref{sec:phase}). As the asteroid rotates, each surface element is lit from
several solar azimuths, much as in rotational photometric stereo, which directly constrains its local normal. Under
near-equatorial station-keeping geometries, however, the body-fixed solar directions are nearly coplanar, so shading
constrains longitudinal slopes well but polar meridional slopes only weakly.

For images with $\alpha\geq 15^\circ$, cast shadows from boulders, crater rims, and concavities enter $V$ through
differentiable shadow mapping \citep{Williams1978}. At each step, the surfels are rendered along $\vect{s}$ from a
virtual Sun camera at 50 body radii to obtain a median-depth map $D_\odot$ of the sunlit body. A surface point
$\vect{x}_p$ at pixel $p$, with depth $z_\odot(\vect{x}_p)$ along the Sun axis and projection $\pi_\odot(\vect{x}_p)$
into the Sun view, has soft visibility
\begin{equation}
V_p = \sigma\!\left(\frac{D_\odot\!\left(\pi_\odot(\vect{x}_p)\right) + b_p - z_\odot(\vect{x}_p)}{\tau}\right),
\label{eq:shadow}
\end{equation}
with slope-dependent bias
\begin{equation}
b_p = b_0 + \min\!\left(\Delta_\odot \tan \theta_p,\ 10\,\Delta_\odot\right),
\label{eq:bias}
\end{equation}
where $\sigma(\cdot)$ is the logistic sigmoid, $\tau$ sets the softness of the shadow edge,
$\theta_p = \arccos(\mzero)$ is evaluated without back-propagating gradients into $\vect{n}$ (preventing noisy normal
gradients from distorting the self-shadow bias), and $\Delta_\odot$ is the pixel scale of the Sun camera. The bias
prevents grazing surfaces from shadowing themselves, while detaching the Sun-view occluder depth map $D_\odot$
(stop-gradient) prevents grazing-shadow extrusion spikes on occluders while preserving forward cast shadows and depth
gradients through $z_\odot(\vect{x}_p)$. To prevent depth noise from creating spurious shadows, $V_p$ is applied
only where the observed brightness falls below a fraction $\gamma$ of the unshadowed prediction $A\,\Phi$ (stopping
gradients across this binary gate for stability), and is set to unity elsewhere. Because this gate requires the
observed image, held-out test views are rendered unshadowed with $V_p=1$.

\subsection{Night-side band and phase switch}
\label{sec:priors}

The silhouette loss $\mathcal{L}_m$ is the binary cross-entropy between the accumulated opacity $O$ and the foreground
mask $M$. At moderate to high phase angles, unlit terrain beyond the terminator is missing from $M$, so an unmodified
silhouette loss carves away the dark hemisphere. We therefore set the silhouette weight to zero inside a directional
night-side band outside $M$ of width
\begin{equation}
w_\mathrm{night} = 1.5\,r\,(1-\cos\alpha) + 5~\mathrm{pixels},
\label{eq:night}
\end{equation}
equal to 1.5 times the unlit crescent width of a sphere of apparent radius $r$ plus a 5-pixel margin. The $1.5$ factor
and 5-pixel margin accommodate non-spherical asteroid elongation, concavities, and limb topography (such as Itokawa's
lobes and Ryugu's or Bennu's equatorial ridges) that extend beyond an inscribed sphere. Shifting $M$ anti-sunward by
$1$ to $\lceil w_\mathrm{night}\rceil$ pixels sweeps out this unlit crescent, disabling the silhouette loss on the dark
side while keeping the sunlit limb supervised. The first term is 2\% of $r$ at $\alpha=10^\circ$ and 35\% at
$40^\circ$. The photometric losses ($\mathcal{L}_1$ and $\mathcal{L}_\mathrm{D\text{-}SSIM}$) stay active inside the
band, and terrain that is dark in one image remains constrained by other views in which it is sunlit.

At low phase angles ($\alpha<15^\circ$), cast shadows are short and shading contrast is weak, making albedo and
surface slope harder to separate. Below $15^\circ$, we turn off shadow mapping and regularise the opacity-normalised
albedo $\tilde A = A/O$ with an isotropic total-variation (TV) penalty:
\begin{equation}
\mathcal{L}_\mathrm{tv} = \frac{1}{|\Omega|}\sum_{p\in\Omega}\Big(\left|\tilde A_{p+\vect{e}_x}-\tilde A_p\right|
 + \left|\tilde A_{p+\vect{e}_y}-\tilde A_p\right|\Big),
\label{eq:tv}
\end{equation}
over foreground pixels $\Omega$ with $O>0.5$, where $\vect{e}_x$ and $\vect{e}_y$ are unit pixel steps. Favouring a
piecewise-smooth albedo attributes gradual brightness changes to surface orientation rather than albedo variations.
We call this choice between the albedo TV prior ($\alpha<15^\circ$) and cast shadows ($\alpha\geq 15^\circ$) the
phase switch; the $15^\circ$ threshold is heuristic (Sect.~\ref{sec:rule}).

\subsection{Loss function}
\label{sec:losses}

Shading constrains surface orientation but not absolute depth. To anchor metric depth, we follow PGSR
\citep{Chen2024PGSR} and compare $7\times 7$ greyscale patches between training view $r$ and a neighbouring view $n$
through the plane-induced homography
\begin{equation}
\vect{H} = \vect{K}_n\left(\vect{R}_{rn} - \frac{\vect{t}_{rn}\vect{n}^\top}{\delta}\right)\vect{K}_r^{-1},
\label{eq:homography}
\end{equation}
where $\vect{K}_r$ and $\vect{K}_n$ are the camera intrinsics, $(\vect{R}_{rn},\vect{t}_{rn})$ is the relative pose,
and $\vect{n}$ and $\delta>0$ are the rendered surface normal and perpendicular plane distance from camera $r$. Patches
that face away from view $n$, fall outside the image, or fail a 2~m multi-view depth-consistency (occlusion) check
after back-projection between views $r$ and $n$ are discarded. The loss $\mathcal{L}_\mathrm{ncc}$ is $1$ minus the
zero-mean normalised cross-correlation (NCC), which is invariant to linear exposure changes. Multi-view NCC thus
constrains depth across textured boulders and crater rims, whereas shading constrains smooth regolith slopes between
tie points, much like stereo and photoclinometry in SPC \citep{Gaskell2008,Palmer2022}. The total loss is
\begin{equation}
\mathcal{L} = (1-\lambda_\mathrm{s})\,\mathcal{L}_1 + \lambda_\mathrm{s}\,\mathcal{L}_\mathrm{D\text{-}SSIM}
 + \lambda_n\mathcal{L}_n + \lambda_m\mathcal{L}_m + \lambda_v\mathcal{L}_\mathrm{ncc} + \lambda_\mathrm{tv}\mathcal{L}_\mathrm{tv},
\label{eq:loss}
\end{equation}
where $\mathcal{L}_1$ and $\mathcal{L}_\mathrm{D\text{-}SSIM}$ measure image residuals \citep{Kerbl2023},
$\mathcal{L}_n$ is the 2DGS normal--depth consistency term \citep{Huang2024}, and $\lambda_\mathrm{tv}=0$ for
$\alpha\geq 15^\circ$ (Table~\ref{tab:settings}).

\subsection{Surface mesh extraction}
\label{sec:extraction}

To remove floating surfels, we keep a foreground pixel ($O>0.5$, $M=1$) only if at least three of six neighbouring
views observe its back-projected point with $O>0.5$ within 1.5~m of their own median depth; on wide-phase sequences
such as Bennu, depth consensus is further restricted to sunlit pixels ($\mzero>0$) so that poorly constrained
dark-limb surfels do not inject outward artefacts into the reconstructed mesh. Points passing this
check on a two-pixel grid are meshed by screened Poisson reconstruction \citep{Kazhdan2013}, keeping the largest
connected component. Because Poisson surfaces tend to bow inwards across poorly observed concavities such as
Itokawa's neck, we also fuse the same depth maps into a truncated signed-distance field \citep[TSDF;][]{Curless1996}
and replace vertices more than 2~m from the TSDF surface (with a two-ring transition band) using discrete biharmonic
fairing \citep{Botsch2004} with fixed boundaries. This combination guarantees a closed, watertight manifold surface
suitable for exact enclosed-volume integration, while vertices modified by fairing represent smoothly interpolated,
unmeasured geometry that should be flagged in operational products. Finally, because rendered median depths lie
$0.1$--$0.2$~pixels in front of the SfM tie points (Sect.~\ref{sec:poses}), we translate the mesh along its
area-weighted normals by the median signed distance of tie points with reprojection residuals $\leq 1$~pixel. None of
these extraction steps uses a reference model.

\section{Data and experimental setup}
\label{sec:setup}

\begin{table*}[!tbp]
\caption{Summary of the five spacecraft image sets.}
\label{tab:data}
\centering
\small\setlength{\tabcolsep}{7.80pt}
\begin{tabular}{@{}llllcccc@{}}
\toprule
Body & Camera & Observation dates & Images & Distance (km) & Pixel scale (m) & Phase angle ($^\circ$) & Splits used \\
\midrule
Ryugu   & Hayabusa2 ONC-T   & 2019 Jan 15         & 60 & 20.18--20.26 & 2.18--2.19 & 7.1--7.3   & $1 \times 50/10$ \\
Itokawa & Hayabusa AMICA    & 2005 Sep 29--Oct 2  & 60 & 7.32--8.00   & 0.73--0.79 & 9.3--10.4  & $3 \times 50/10$ \\
Ryugu   & Hayabusa2 ONC-T   & 2018 Jul 10         & 60 & 19.90--20.09 & 2.15--2.17 & 18.5--18.7 & $3 \times 50/10$ \\
Ryugu   & Hayabusa2 ONC-T   & 2018 Aug 31         & 48 & 21.77--21.89 & 2.35--2.36 & 39.9--40.3 & $2 \times 40/8$ \\
Bennu   & OSIRIS-REx MapCam & 2018 Nov 30--Dec 13 & 60 & 9.07--38.31  & 0.62--2.60 & 26.1--59.9 & $3 \times 50/10$ \\
\bottomrule
\end{tabular}
\end{table*}

\begin{figure*}[!tbp]
\centering
\includegraphics[width=\textwidth]{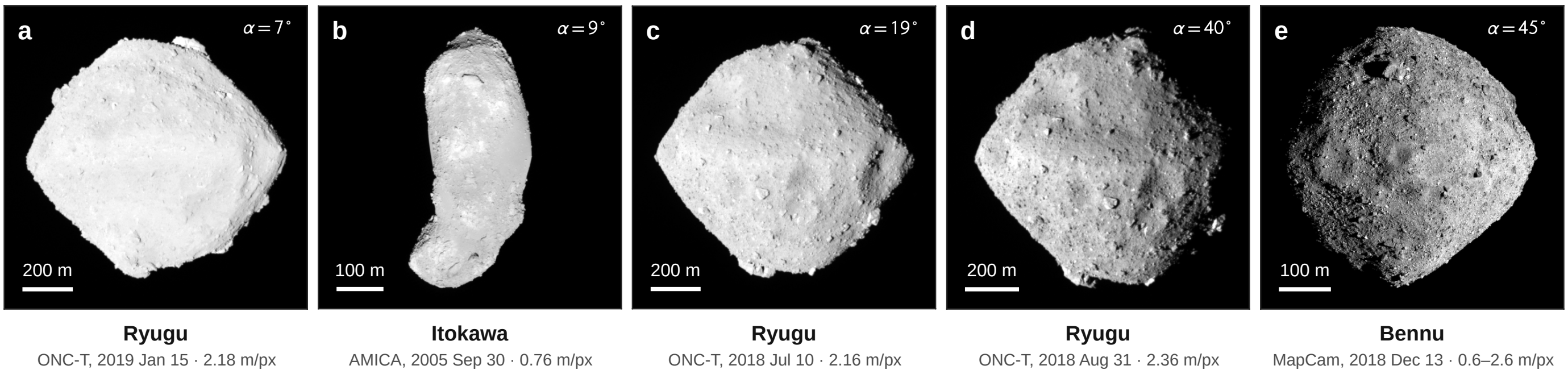}
\caption{Representative test views from the five campaigns ordered by solar phase angle $\alpha$, annotated with
ground pixel scales (0.6--2.6~m per pixel): (a) Ryugu (ONC-T, 2019 January 15, $\alpha=7^\circ$, 2.18~m per pixel);
(b) Itokawa (AMICA, 2005 September 30, $\alpha=9^\circ$, 0.76~m per pixel); (c) Ryugu (ONC-T, 2018 July 10,
$\alpha=19^\circ$, 2.16~m per pixel); (d) Ryugu (ONC-T, 2018 August 31, $\alpha=40^\circ$, 2.36~m per pixel); and
(e) Bennu (MapCam, 2018 December 13, $\alpha=45^\circ$, 0.6--2.6~m per pixel across the sequence). Higher phase
angles (d, e) leave extensive unlit terrain beyond the terminator.}
\label{fig:campaigns}
\end{figure*}

\subsection{Image sets}
\label{sec:images}

We evaluated \pcgs\ on five public spacecraft image sets of three near-Earth asteroids (Table~\ref{tab:data};
Fig.~\ref{fig:campaigns}). Itokawa was imaged by the Asteroid Multi-band Imaging Camera (AMICA) on board Hayabusa from
ranges of 7.3--8.0~km \citep{Ishiguro2010,Saito2006}. Ryugu was observed by the telescopic Optical Navigation Camera
(ONC-T) on board Hayabusa2 from near the 20~km Box-A station-keeping altitude during three hovering periods at phase
angles of $\alpha\approx 7^\circ$, $19^\circ$, and $40^\circ$ \citep{Kameda2017,Watanabe2019,ChenH2025}, providing
views of the same body under three illumination geometries. Bennu was imaged by MapCam on board OSIRIS-REx during the
Approach and Preliminary Survey phases from ranges of 9.1--38.3~km and at phase angles of $26^\circ$--$60^\circ$
\citep{Rizk2018}. Across these campaigns, sub-spacecraft latitudes lie within $\pm 5^\circ$ on Itokawa and Ryugu and
within $[-4^\circ, +13^\circ]$ on Bennu, while sub-solar latitudes lie within $[-8^\circ, +3^\circ]$.

For Itokawa and Ryugu, we selected 60 full-disc frames evenly spaced in sub-spacecraft longitude (all 48 available
frames for Ryugu at $40^\circ$). For Bennu, 60 frames were selected from SPICE geometry alone, so a few close-range
views exhibit minor limb cropping. All images have a size of $1024\times 1024$~pixels. The Hayabusa2 ONC-T (v-band,
Level-2c) and OSIRIS-REx MapCam (panchromatic, Level-2 $I/F$) archival products are stored in floating-point format; we
linearly scaled each dataset to 8-bit greyscale using a single scale factor (the 99.9th percentile of positive pixel
values) so that relative brightness across frames is preserved. The Hayabusa AMICA v-band frames were used directly in
their archived 8-bit digital-number (DN) format without per-frame exposure normalisation. Each dataset was divided three
times at random into 50 training and 10 test images (40 and 8 for Ryugu at $40^\circ$). We evaluated all three splits
on Itokawa, Ryugu ($19^\circ$), and Bennu, two splits on Ryugu ($40^\circ$), and one split on Ryugu ($7^\circ$).

\subsection{Camera poses, masks, and reference models}
\label{sec:poses}
\label{sec:reference}

Initial camera poses and solar vectors were computed from the mission SPICE kernels \citep{Acton1996}. Because the
narrow fields of view of spacecraft framing cameras can cause unconstrained structure-from-motion (SfM) to fail or
converge to a depth-inverted solution, we initialised COLMAP \citep{Schonberger2016} with these SPICE poses, matched
SIFT features across all image pairs, and refined the camera extrinsics with fixed pinhole intrinsics (focal lengths of
$10\,066.7$, $9269.2$, and $14\,729.4$~pixels for AMICA, ONC-T, and MapCam) while tightening the reprojection threshold
from 40 to 4~pixels. All campaign frames entered the bundle adjustment, reflecting operational optical navigation, and the identical refined
poses, tie points, post-processing alignment, and evaluation registration were shared across all methods. Because no
test frame optimised the surfels and the scalar normal offset aggregates thousands of tie points, test-frame influence
is minimal, though retaining held-out frames in bundle adjustment remains a minor caveat. A seven-parameter similarity
transform aligned the refined camera centres to the SPICE trajectory in the metric body-fixed frame so that SPICE solar
vectors apply directly. Sunlit foreground masks threshold each image at 8~DN (AMICA) or 4\% of the dataset
99.9th-percentile intensity (ONC-T and MapCam), keeping the largest filled component and refining its outer 3-pixel
border at 50\% local limb brightness.

Reconstructions of Itokawa and Ryugu were evaluated against the 3.1-million-facet SPC model \citep{Gaskell2008} and the
3.0-million-facet SfM-MVS model \citep{Watanabe2019}, respectively. Bennu was evaluated against the image-independent
3.15-million-facet OLA v21 laser-altimeter model \citep[0.88~m mean edge length;][]{Seabrook2022}. To remove the small
coordinate-frame and scale offset between the SPICE-anchored camera frame and each reference (scale factors
0.997--1.011; Table~\ref{tab:volume}), we fitted a single evaluation-only similarity transform from the sparse SfM tie
points to the reference surface using trimmed iterative closest point (ICP) registration and applied it identically to
all methods.

\subsection{Evaluation metrics and post-processing}
\label{sec:metrics}

Following \citet{Zhang2025AsteroidGS}, we report the mean and root mean square error (RMSE) of the unsigned distance
from each reconstructed vertex to the reference surface. Because one-sided vertex distances depend on mesh tessellation
and do not penalise holes, we also sample $N=10^6$ points uniformly by area on the reconstructed ($\mathcal{X}$) and
reference ($\mathcal{Y}$) surfaces and compute the accuracy $P(\epsilon)$ and completeness $R(\epsilon)$:
\begin{equation}
\begin{aligned}
P(\epsilon) &= \frac{1}{N}\sum_{\vect{x}\in\mathcal{X}}\left[d(\vect{x},\mathcal{Y})<\epsilon\right], \\
R(\epsilon) &= \frac{1}{N}\sum_{\vect{y}\in\mathcal{Y}}\left[d(\vect{y},\mathcal{X})<\epsilon\right],
\end{aligned}
\label{eq:pr}
\end{equation}
where $d(\cdot,\cdot)$ denotes the unsigned point-to-surface distance and $[\cdot]$ is the Iverson bracket ($1$ if
true, $0$ otherwise). Their harmonic mean yields the F-score $2PR/(P+R)$ at $\epsilon=1$~m. From the same sample sets,
we also compute the mean completeness distance $d(\vect{y},\mathcal{X})$, the symmetric Chamfer distance (the average
of the two directed mean distances), and the normal error (the mean acute angle between facet normals at
$\vect{x}\in\mathcal{X}$ and those of the nearest reference facets). For novel-view synthesis (NVS) on held-out test
images, we report the peak signal-to-noise ratio (PSNR), the structural similarity index \citep[SSIM;][]{Wang2004SSIM},
and the learned perceptual image patch similarity \citep[LPIPS with a VGG backbone;][]{Zhang2018LPIPS}. Because methods
with per-frame appearance codes have no fitted code for unseen test views, these metrics are computed after fitting a
linear gain and offset per image over the foreground mask.

\begin{table}[!htbp]
\centering
\caption{Mean vertex-to-surface distance (m) and F-score at 1~m across post-processing protocols (three-split means).}
\label{tab:protocol}
\scriptsize
\setlength{\tabcolsep}{4.18pt}
\begin{tabular}{lrrrrrrrr}
\toprule
 & \multicolumn{2}{c}{Extracted} & \multicolumn{2}{c}{+ tie-point} & \multicolumn{2}{c}{+ Taubin} & \multicolumn{2}{c}{ICP reg.} \\
\cmidrule(lr){2-3}\cmidrule(lr){4-5}\cmidrule(lr){6-7}\cmidrule(lr){8-9}
Method & Mean & F@1 & Mean & F@1 & Mean & F@1 & Mean & F@1 \\
\midrule
\multicolumn{9}{l}{\textit{Itokawa (AMICA, phase angle 10$^\circ$)}} \\
2DGS & 6.58 & 0.532 & 6.55 & 0.540 & 6.55 & 0.546 & 6.56 & 0.539 \\
GOF & 2.41 & 0.365 & 2.40 & 0.365 & 2.40 & 0.362 & 2.39 & 0.373 \\
PGSR & \underline{1.18} & \underline{0.695} & 0.99 & \underline{0.811} & 0.99 & \underline{0.812} & 0.98 & \underline{0.821} \\
Asteroid-GS & 1.52 & 0.432 & 1.01 & 0.648 & 1.01 & 0.649 & 0.99 & 0.661 \\
Asteroid-GS$^{\dagger}$ & 1.53 & 0.415 & \underline{0.94} & 0.659 & \underline{0.94} & 0.658 & \underline{0.91} & 0.684 \\
\rowcolor{pcgsrow} \textbf{PCGS (this work)} & \textbf{0.59} & \textbf{0.853} & \textbf{0.59} & \textbf{0.854} & \textbf{0.59} & \textbf{0.853} & \textbf{0.59} & \textbf{0.857} \\
\addlinespace[1.5pt]
\multicolumn{9}{l}{\textit{Ryugu (ONC-T, phase angle 19$^\circ$)}} \\
2DGS & 2.51 & \underline{0.388} & 2.51 & 0.410 & 2.50 & 0.416 & 2.48 & 0.421 \\
GOF & 4.20 & 0.156 & 4.21 & 0.177 & 4.20 & 0.162 & 3.38 & 0.255 \\
PGSR & \underline{1.51} & 0.383 & \underline{0.73} & \underline{0.788} & \underline{0.73} & \underline{0.787} & \underline{0.72} & \underline{0.794} \\
Asteroid-GS & 2.41 & 0.239 & 1.27 & 0.524 & 1.27 & 0.525 & 1.26 & 0.530 \\
Asteroid-GS$^{\dagger}$ & 2.43 & 0.242 & 1.29 & 0.529 & 1.29 & 0.530 & 1.27 & 0.539 \\
\rowcolor{pcgsrow} \textbf{PCGS (this work)} & \textbf{0.62} & \textbf{0.797} & \textbf{0.62} & \textbf{0.798} & \textbf{0.62} & \textbf{0.798} & \textbf{0.61} & \textbf{0.803} \\
\bottomrule
\end{tabular}
\tablefoot{$^{\dagger}$Default camera-ring scene extent.}
\end{table}

All reconstructed meshes underwent the same reference-free post-processing: a global translation and scalar normal
offset fitted to the SfM tie points to remove global frame and depth biases, 20 iterations of volume-preserving Taubin
smoothing \citep{Taubin1995}, and selection of the largest connected component. As verified in
Table~\ref{tab:protocol}, method rankings remain unchanged across the raw, tie-aligned, smoothed, and per-mesh ICP
protocols.

\begin{table}[!htbp]
\caption{Hyperparameters and configuration of \pcgs, held fixed across all five image sets.}
\label{tab:settings}
\centering
\footnotesize\setlength{\tabcolsep}{4.8pt}
\begin{tabular}{@{}ll@{}}
\toprule
Component & Configuration \\
\midrule
Optimisation & $30\,000$ iterations; 2DGS Adam schedule \\
Appearance & Relative albedo; $L$ learned (init.\ $0.9$, rate $10^{-2}$) \\
Shading warm-up & Unshaded $\leq$ iter.\ 3000; shaded $>$ iter.\ 3000 \\
Photometric loss & $\lambda_\mathrm{s}=0.2$ (D-SSIM weight) \\
Normal consistency & $\lambda_n=0.05$ from iter.\ 7000 \\
Silhouette loss & $\lambda_m=0.05$; night-side band via Eq.~\eqref{eq:night} \\
Multi-view NCC & $\lambda_v=0.15$ from iter.\ 7000; 4096 patches ($7\times7$); \\
 & $\leq$4 neighbours at $2^\circ$--$25^\circ$ separation \\
Albedo TV prior & $\lambda_\mathrm{tv}=0.1$ for $\alpha<15^\circ$ (from iter.\ 3000) \\
Cast shadows & $\alpha\geq15^\circ$; $1024^2$ Sun view (FoV $3^\circ$); \\
 & detached $D_\odot$; $\tau=0.5$~m, $b_0=3$~m, $\gamma=0.9$ \\
Mesh extraction & $\geq$3 of 6 views ($3^\circ$--$40^\circ$) within 1.5~m; \\
 & sunlit on wide-phase; depth 11; fairing $>2$~m \\
Object extent & $1.1\times$ 99th-percentile SfM tie-point radius \\
\bottomrule
\end{tabular}
\end{table}

\subsection{Implementation details and baselines}
\label{sec:implementation}

Table~\ref{tab:settings} summarises the hyperparameters of \pcgs, which were held fixed across all five datasets.
Surfels were initialised at the sparse SfM tie points (after radial-outlier filtering on wide-phase sequences to
remove far-field triangulation noise) and optimised with the 2DGS depth-distortion regulariser disabled
($\lambda_\mathrm{dist}=0$). During the first 3000 iterations, the unshaded scalar albedo $A$ was fitted directly to
the images to establish coarse geometry; lunar-Lambert shading (with logistic-parametrised $L$) and the active
phase-switch branch were enabled thereafter. From iteration 7000, $\mathcal{L}_\mathrm{ncc}$ pairs each training view
at random with up to four neighbouring views separated by $2^\circ$--$25^\circ$. The virtual Sun camera has a
resolution of $1024\times 1024$~pixels and a $3^\circ$ field of view (FoV), and gradients are not back-propagated
through the Sun-view depth map $D_\odot$, the shadow gate, or the incidence angle $\theta_p$. In Eq.~\eqref{eq:night},
the apparent radius $r$ is the 99th-percentile tie-point radius projected into pixels. For mesh extraction, TSDF fusion
uses 1~m voxels and a 4~m truncation distance.

We compared \pcgs\ with six baselines: five surface-oriented Gaussian-splatting methods -- 2DGS \citep{Huang2024}, GOF
\citep{Yu2024GOF}, RaDe-GS \citep{Zhang2024RaDe}, PGSR \citep{Chen2024PGSR}, and Asteroid-GS \citep{Zhang2025AsteroidGS}
-- plus classical COLMAP dense multi-view stereo \citep[MVS;][]{Schonberger2016MVS}. All Gaussian-splatting baselines
used their released code and identical inputs, with object-bounded TSDF extraction for 2DGS and the released
expected-depth TSDF extraction for RaDe-GS. Because Gaussian splatting scales its densification thresholds and position
learning rates by the scene extent, spacecraft distances (${\sim}20$~km at Ryugu) that are much larger than the asteroid
radius (${\sim}0.5$~km) make the default camera-ring extent far too coarse for the target. All Gaussian-splatting
methods therefore used a shared object extent equal to $1.1$ times the 99th-percentile tie-point radius; for
completeness, we also evaluated Asteroid-GS with its default camera-ring extent (denoted Asteroid-GS$^\dagger$). None of
the Gaussian-splatting baselines uses a training silhouette loss: 2DGS, RaDe-GS, and PGSR mask rendered depths only
prior to TSDF fusion, whereas GOF and Asteroid-GS run unmasked. On Bennu, GOF exceeded the 80~GB GPU memory limit
during tetrahedral meshing and could not be completed. For COLMAP dense MVS, we ran PatchMatch stereo depth estimation,
geometric stereo fusion, and depth-10 screened Poisson meshing on the exact same 50 training images (40 for Ryugu at
$40^\circ$) and refined SfM poses, followed by the same reference-free post-processing; COLMAP dense MVS produces
geometry only and does not synthesise novel views. Across the phase-angle sweep (Sect.~\ref{sec:phase}), COLMAP dense
MVS was evaluated on Itokawa, Ryugu ($19^\circ$ and $40^\circ$), and Bennu, while PGSR and Asteroid-GS were additionally
evaluated on Ryugu at $7^\circ$ and $40^\circ$. A TSDF mesh was also fused from volumetric Mip-Splatting
\citep{Yu2024Mip} depths on the first split of Itokawa as an image-fidelity reference.

One implementation detail affects the Gaussian-splatting baseline comparison. Unmodified 2DGS-derived codes mix
inconsistent pixel-centre and principal-point conventions across stages: rasterisation assumes a principal point at
$(W-1)/2$ for an image of width $W$, COLMAP places the centre of pixel $i$ at $i+1/2$, and depth back-projection assumes
$W/2$, shifting the reconstructed surface by roughly 1~pixel relative to the bundle-adjusted cameras. Aligning both
rasterisation and depth back-projection to COLMAP in \pcgs\ reduces depth reprojection residuals to $10^{-3}$~pixels.
For the unmodified baselines, the tie-point alignment removes the global rigid component of this shift, amounting to
$2.15$~m for PGSR and $1.9$~m for Asteroid-GS on Ryugu at $19^\circ$, which reduces PGSR's mean distance from $1.51$ to
$0.73$~m (Table~\ref{tab:protocol}). All runs were executed on a single NVIDIA A100 (80~GB) GPU. \pcgs\ required
17--47~min for training (median 27~min) plus 4--10~min for mesh extraction, comparable to PGSR (29--31~min) and
Asteroid-GS (22--52~min) on Itokawa and Ryugu.

\begin{table*}[!tbp]
\centering
\caption{Shape reconstruction and novel-view synthesis accuracy on Itokawa, Ryugu, and Bennu (mean $\pm$ standard deviation over three splits of 50 training and 10 test images).}
\label{tab:main}
\footnotesize
\setlength{\tabcolsep}{11.30pt}
\begin{tabular}{lcccccc}
\toprule
 & \multicolumn{5}{c}{Shape accuracy relative to reference model} & Novel views \\
\cmidrule(lr){2-6}\cmidrule(lr){7-7}
Method & Mean (m) & RMSE (m) & Chamfer (m) & F@1~m & Normal ($^\circ$) & PSNR (dB) \\
\midrule
\multicolumn{7}{l}{\textit{Itokawa (AMICA, phase angle 10$^\circ$; SPC reference)}} \\
COLMAP MVS & 0.85\,$\pm$\,0.01 & 1.49\,$\pm$\,0.04 & 0.71\,$\pm$\,0.01 & 0.808\,$\pm$\,0.003 & 15.1\,$\pm$\,0.1 & -- \\
2DGS & 6.55\,$\pm$\,3.41 & 17.96\,$\pm$\,11.62 & 4.63\,$\pm$\,1.92 & 0.546\,$\pm$\,0.035 & 24.1\,$\pm$\,2.4 & 33.6\,$\pm$\,1.0 \\
GOF & 2.40\,$\pm$\,0.12 & 4.66\,$\pm$\,0.45 & 3.01\,$\pm$\,0.24 & 0.362\,$\pm$\,0.004 & 41.5\,$\pm$\,0.4 & 38.0\,$\pm$\,0.6 \\
RaDe-GS & \underline{0.62}\,$\pm$\,0.01 & \textbf{0.94}\,$\pm$\,0.03 & \underline{0.66}\,$\pm$\,0.00 & \underline{0.828}\,$\pm$\,0.003 & \underline{13.6}\,$\pm$\,0.2 & \textbf{40.0}\,$\pm$\,0.9 \\
PGSR & 0.99\,$\pm$\,0.07 & 2.83\,$\pm$\,0.46 & 0.76\,$\pm$\,0.03 & 0.812\,$\pm$\,0.001 & 14.3\,$\pm$\,0.1 & \underline{39.3}\,$\pm$\,0.8 \\
Asteroid-GS & 1.01\,$\pm$\,0.10 & 2.10\,$\pm$\,0.41 & 1.76\,$\pm$\,0.55 & 0.649\,$\pm$\,0.031 & 13.9\,$\pm$\,1.1 & 29.6\,$\pm$\,0.9 \\
Asteroid-GS$^{\dagger}$ & 0.94\,$\pm$\,0.03 & 1.82\,$\pm$\,0.17 & 1.54\,$\pm$\,0.16 & 0.658\,$\pm$\,0.008 & \textbf{12.7}\,$\pm$\,0.2 & 29.5\,$\pm$\,0.5 \\
\rowcolor{pcgsrow} \textbf{PCGS (this work)} & \textbf{0.59}\,$\pm$\,0.02 & \underline{1.02}\,$\pm$\,0.15 & \textbf{0.59}\,$\pm$\,0.01 & \textbf{0.853}\,$\pm$\,0.005 & 14.7\,$\pm$\,0.2 & 36.5\,$\pm$\,0.5 \\
\addlinespace[2pt]
\multicolumn{7}{l}{\textit{Ryugu (ONC-T, phase angle 19$^\circ$; SfM-MVS reference)}} \\
COLMAP MVS & 1.13\,$\pm$\,0.01 & 1.76\,$\pm$\,0.04 & 0.87\,$\pm$\,0.01 & 0.718\,$\pm$\,0.003 & 11.9\,$\pm$\,0.1 & -- \\
2DGS & 2.50\,$\pm$\,0.08 & 3.80\,$\pm$\,0.17 & 1.96\,$\pm$\,0.07 & 0.416\,$\pm$\,0.010 & 28.0\,$\pm$\,0.4 & 34.9\,$\pm$\,1.8 \\
GOF & 4.20\,$\pm$\,0.08 & 5.98\,$\pm$\,0.15 & 4.50\,$\pm$\,0.11 & 0.162\,$\pm$\,0.003 & 42.4\,$\pm$\,0.3 & 33.5\,$\pm$\,0.0 \\
RaDe-GS & 3.12\,$\pm$\,0.85 & 4.19\,$\pm$\,1.23 & 3.43\,$\pm$\,0.86 & 0.249\,$\pm$\,0.050 & 32.6\,$\pm$\,7.7 & \textbf{35.8}\,$\pm$\,0.9 \\
PGSR & \underline{0.73}\,$\pm$\,0.01 & \underline{1.18}\,$\pm$\,0.04 & \underline{0.76}\,$\pm$\,0.01 & \underline{0.787}\,$\pm$\,0.005 & 14.5\,$\pm$\,0.1 & \underline{35.2}\,$\pm$\,0.2 \\
Asteroid-GS & 1.27\,$\pm$\,0.02 & 1.88\,$\pm$\,0.03 & 1.42\,$\pm$\,0.03 & 0.525\,$\pm$\,0.005 & \underline{11.4}\,$\pm$\,0.0 & 22.7\,$\pm$\,0.2 \\
Asteroid-GS$^{\dagger}$ & 1.29\,$\pm$\,0.01 & 1.98\,$\pm$\,0.02 & 1.52\,$\pm$\,0.02 & 0.530\,$\pm$\,0.004 & \textbf{11.3}\,$\pm$\,0.0 & 22.7\,$\pm$\,0.0 \\
\rowcolor{pcgsrow} \textbf{PCGS (this work)} & \textbf{0.62}\,$\pm$\,0.00 & \textbf{0.89}\,$\pm$\,0.01 & \textbf{0.73}\,$\pm$\,0.00 & \textbf{0.798}\,$\pm$\,0.000 & 12.1\,$\pm$\,0.0 & 31.3\,$\pm$\,0.1 \\
\addlinespace[2pt]
\multicolumn{7}{l}{\textit{Bennu (MapCam, phase angle 26--60$^\circ$; OLA laser-altimeter reference)}} \\
COLMAP MVS & \underline{0.53}\,$\pm$\,0.01 & \textbf{0.71}\,$\pm$\,0.05 & \textbf{0.67}\,$\pm$\,0.07 & \textbf{0.818}\,$\pm$\,0.021 & \textbf{29.1}\,$\pm$\,0.3 & -- \\
2DGS & 26.77\,$\pm$\,1.59 & 41.64\,$\pm$\,2.69 & 26.85\,$\pm$\,0.33 & 0.138\,$\pm$\,0.023 & 53.3\,$\pm$\,0.8 & 20.1\,$\pm$\,0.6 \\
RaDe-GS & 1.21\,$\pm$\,0.33 & 4.76\,$\pm$\,1.72 & 3.72\,$\pm$\,1.58 & 0.760\,$\pm$\,0.038 & 32.3\,$\pm$\,0.6 & \underline{20.7}\,$\pm$\,0.3 \\
PGSR & 3.78\,$\pm$\,0.85 & 11.26\,$\pm$\,3.39 & 3.84\,$\pm$\,0.81 & 0.689\,$\pm$\,0.049 & 38.0\,$\pm$\,1.3 & 20.3\,$\pm$\,0.3 \\
Asteroid-GS$^{\dagger}$ & 3.78\,$\pm$\,1.75 & 8.69\,$\pm$\,5.00 & 5.94\,$\pm$\,3.47 & 0.368\,$\pm$\,0.035 & 33.1\,$\pm$\,1.3 & 17.6\,$\pm$\,1.0 \\
\rowcolor{pcgsrow} \textbf{PCGS (this work)} & \textbf{0.50}\,$\pm$\,0.01 & \underline{0.78}\,$\pm$\,0.02 & \underline{0.76}\,$\pm$\,0.01 & \underline{0.794}\,$\pm$\,0.010 & \underline{31.6}\,$\pm$\,0.3 & \textbf{22.7}\,$\pm$\,0.6 \\
\bottomrule
\end{tabular}
\tablefoot{Bold and underlined values mark the best and second-best means, respectively; PSNR is gain- and offset-matched. COLMAP MVS (PatchMatch stereo with depth-10 screened Poisson meshing) does not synthesise novel views (marked --). RaDe-GS uses its released expected-depth TSDF extraction. $^{\dagger}$Asteroid-GS with its default camera-ring scene extent (all other Gaussian-splatting rows use the shared object extent).}
\end{table*}

\begin{figure*}[!tbp]
\centering
\includegraphics[width=\textwidth]{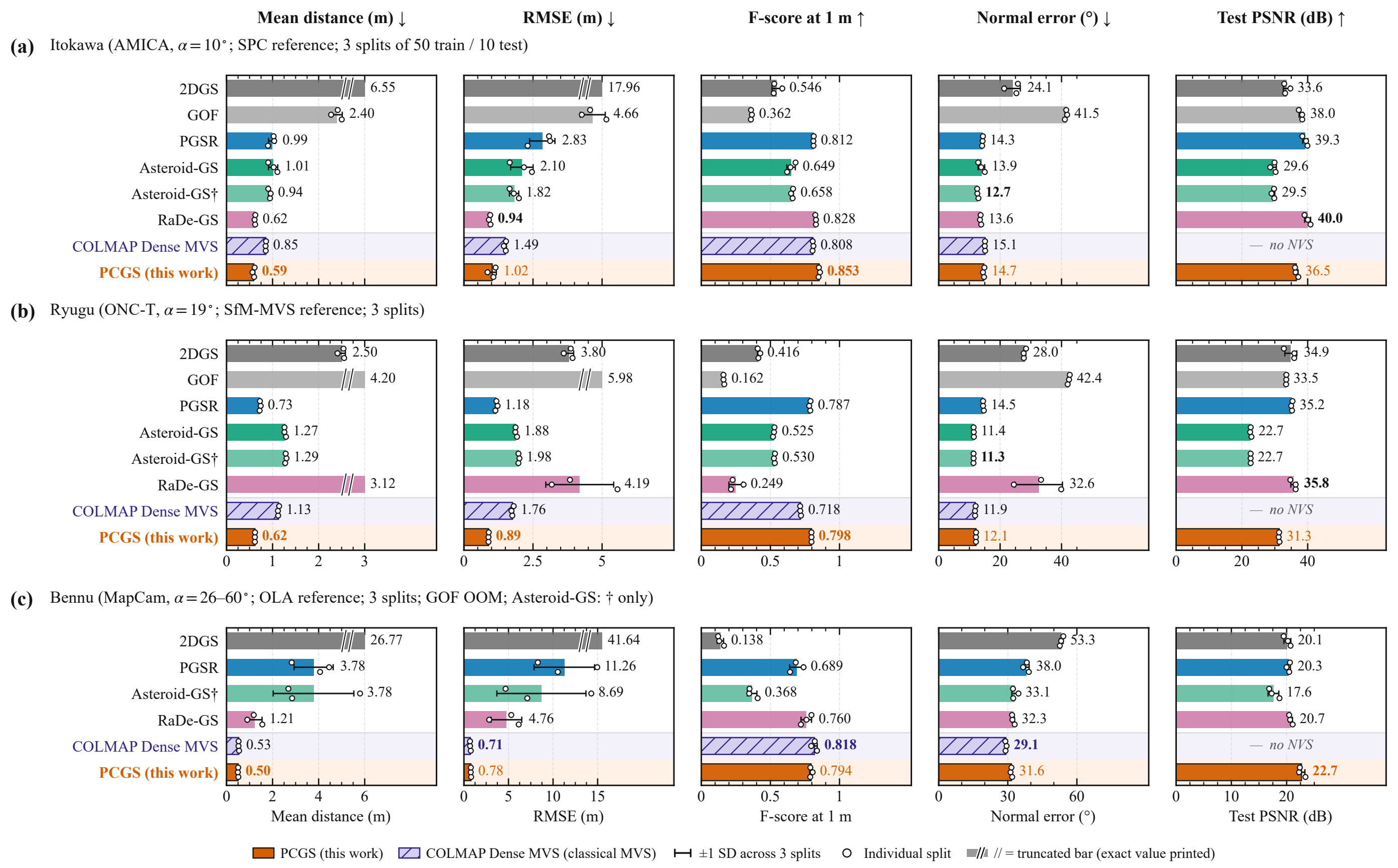}
\caption{Benchmark comparison of shape and novel-view synthesis accuracy across all evaluated methods on (a) Itokawa
($\alpha=10^\circ$), (b) Ryugu ($\alpha=19^\circ$), and (c) Bennu ($\alpha=26^\circ$--$60^\circ$, evaluated against OLA
laser altimetry). Horizontal bars show three-split means with $\pm 1$ standard deviation error bars and open circles
for individual splits; bars exceeding the panel scale are truncated with double slashes (\texttt{//}) and labelled with
their exact values, and bold numerals mark the best mean per column. COLMAP dense MVS is shown with a hatched bar and
omits novel-view PSNR (no NVS). $^{\dagger}$Asteroid-GS with camera-ring scene extent.}
\label{fig:numbers}
\end{figure*}

\section{Results}
\label{sec:results}

\subsection{Global shape accuracy and enclosed volumes}
\label{sec:accuracy}

Table~\ref{tab:main} and Fig.~\ref{fig:numbers}a,b compare three-split reconstruction accuracy on Itokawa
($\alpha=10^\circ$) and Ryugu ($\alpha=19^\circ$), combining one-sided vertex-to-surface errors (Mean, RMSE) with
area-uniform surface metrics (Chamfer, F@1~m; Sect.~\ref{sec:metrics}) so that neither mesh tessellation nor holes can
mask geometric bias. Because Table~\ref{tab:main} compares complete systems (including silhouette supervision,
terminator masking, and pixel-centre alignment in \pcgs), the controlled leave-one-out ablations in
Sect.~\ref{sec:ablation} isolate the photometric gain from pipeline effects. \pcgs\ ranks first on both bodies in mean
vertex distance ($0.59\pm 0.02$~m on Itokawa; $0.62\pm 0.00$~m on Ryugu), Chamfer distance ($0.59\pm 0.01$ and
$0.73\pm 0.00$~m), and F@1~m ($0.853\pm 0.005$ and $0.798\pm 0.000$), while ranking first in RMSE on Ryugu
($0.89\pm 0.01$~m) and second to RaDe-GS on Itokawa ($1.02\pm 0.15$ vs.\ $0.94\pm 0.03$~m). On Itokawa, \pcgs\ beats
all six baselines in mean, Chamfer, and F@1~m, including RaDe-GS ($0.62$~m mean, $0.66$~m Chamfer, F@1~m $0.828$),
COLMAP dense MVS ($0.85$~m mean, $1.49$~m RMSE; 31\% lower mean error for \pcgs), Asteroid-GS$^\dagger$ ($0.94$~m), and
PGSR ($0.99$~m); moreover, all three RaDe-GS Itokawa meshes are non-watertight ($-4.6\%$ area deficit, undefined
volume). On Ryugu ($19^\circ$), RaDe-GS breaks down ($3.12\pm 0.85$~m mean, $4.19\pm 1.23$~m RMSE, $3.43\pm 0.86$~m
Chamfer, F@1~m $0.249$) alongside 2DGS ($2.50$~m) and GOF ($4.20$~m). Here \pcgs\ lowers mean distance by 15\% vs.\
PGSR ($0.62$ vs.\ $0.73$~m; RMSE down 24\%, $0.89$ vs.\ $1.18$~m), 45\% vs.\ COLMAP dense MVS ($1.13$~m mean, $1.76$~m
RMSE), and fivefold vs.\ RaDe-GS across all protocols (Table~\ref{tab:protocol}). Extended metrics
(Table~\ref{tab:shape_full}; Fig.~\ref{fig:curves}a--d) show the largest lead at sub-metre scales (F@0.5~m $0.579$ vs.\
$0.542$ RaDe-GS, $0.527$ COLMAP MVS, $0.526$ PGSR on Itokawa; $0.516$ vs.\ $0.499$ PGSR and $0.436$ COLMAP MVS on
Ryugu), whereas on Ryugu \pcgs\ and PGSR match closely at coarser scales ($0.73$ vs.\ $0.76$~m Chamfer; F@2~m $0.948$
vs.\ $0.949$), confirming that RMSE gains stem from removing local outliers.

\begin{figure*}[!p]
\centering
\includegraphics[width=166mm]{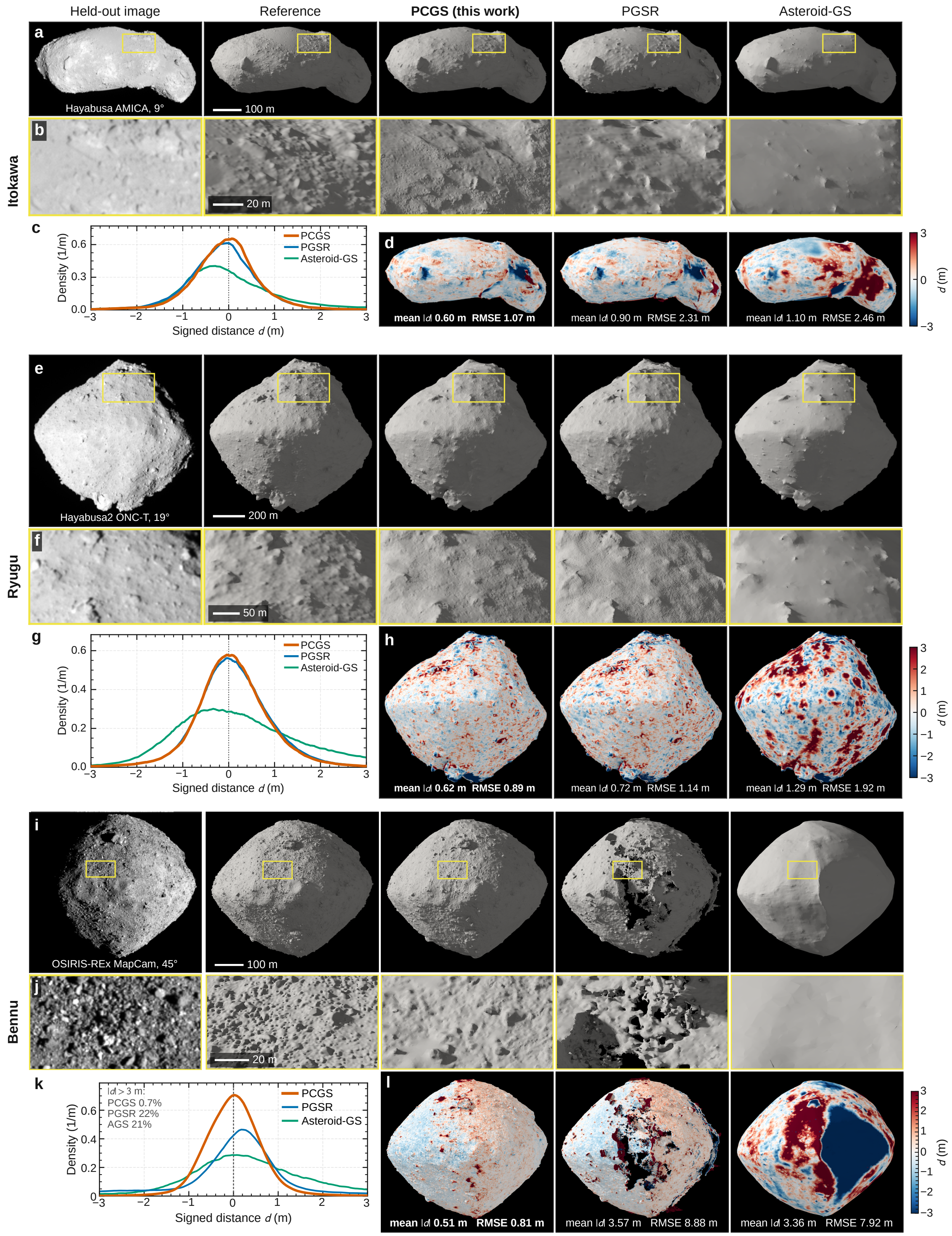}
\caption{Reconstructed global shapes on Itokawa, Ryugu ($\alpha=19^\circ$), and Bennu (first split). (a, e, i) Test
image and Lambertian shaded views of the reference, \pcgs, PGSR, and Asteroid-GS from identical poses (Itokawa
rotated $90^\circ$); yellow boxes mark close-ups. (b, f, j) Boulder close-ups. (c, g, k) Signed distance distributions
$d$ (positive outwards; $|d|>3$~m fractions noted in k). (d, h, l) Signed distance maps ($\pm 3$~m) with mean $|d|$
and RMSE. Asteroid-GS uses object extent on Itokawa and Ryugu and default camera-ring extent on Bennu.}
\label{fig:recon}
\end{figure*}

\begin{figure*}[!tbp]
\centering
\includegraphics[width=\textwidth]{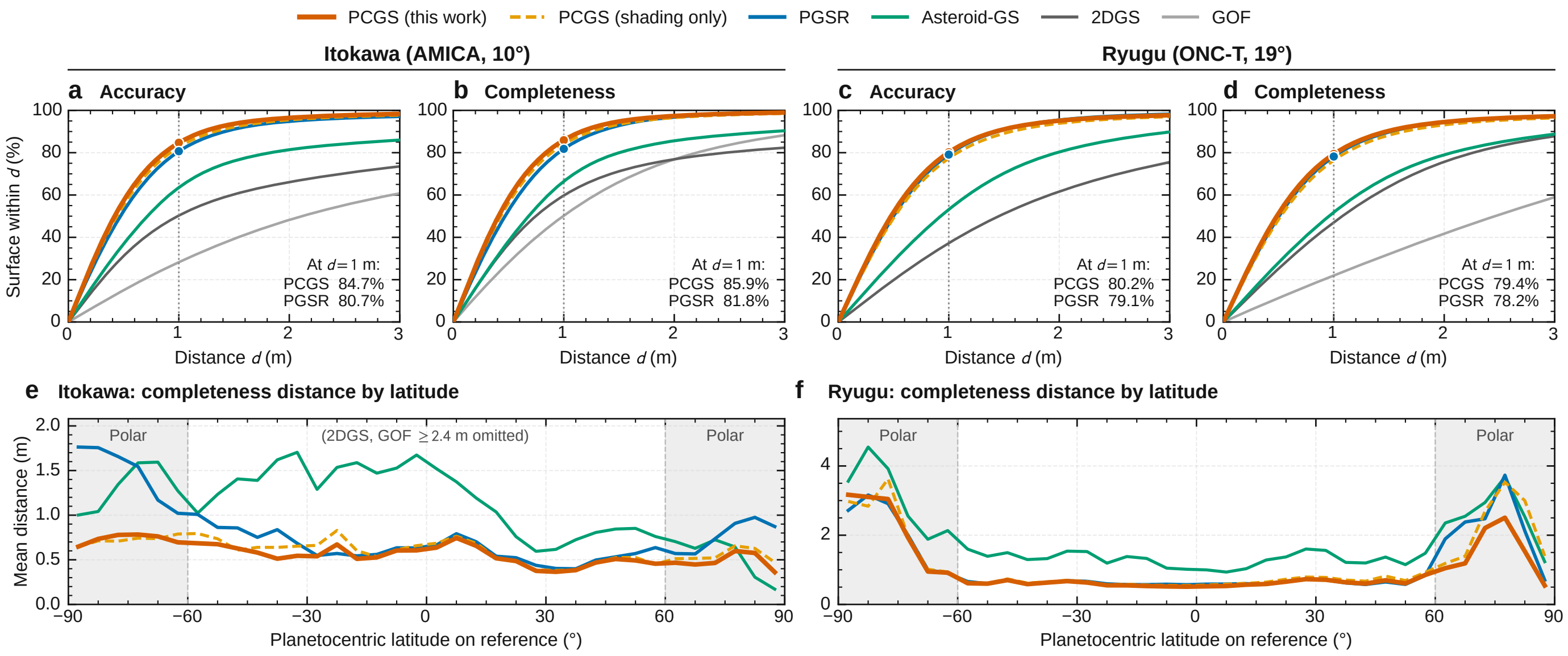}
\caption{Cumulative error distributions and latitudinal completeness profiles on Itokawa and Ryugu
($\alpha=19^\circ$). (a--d) Three-split mean cumulative accuracy and completeness versus distance $d$, annotated at
$d=1$~m for \pcgs\ and PGSR; dashed orange curves show \pcgs\ with shading alone. (e, f) Mean completeness distance in
$5^\circ$ latitude bins (grey: $>60^\circ$ latitude); 2DGS and GOF ($\geq 2.4$~m) are omitted from (e) for clarity.}
\label{fig:curves}
\end{figure*}

\begin{figure*}[!tbp]
\centering
\includegraphics[width=\textwidth]{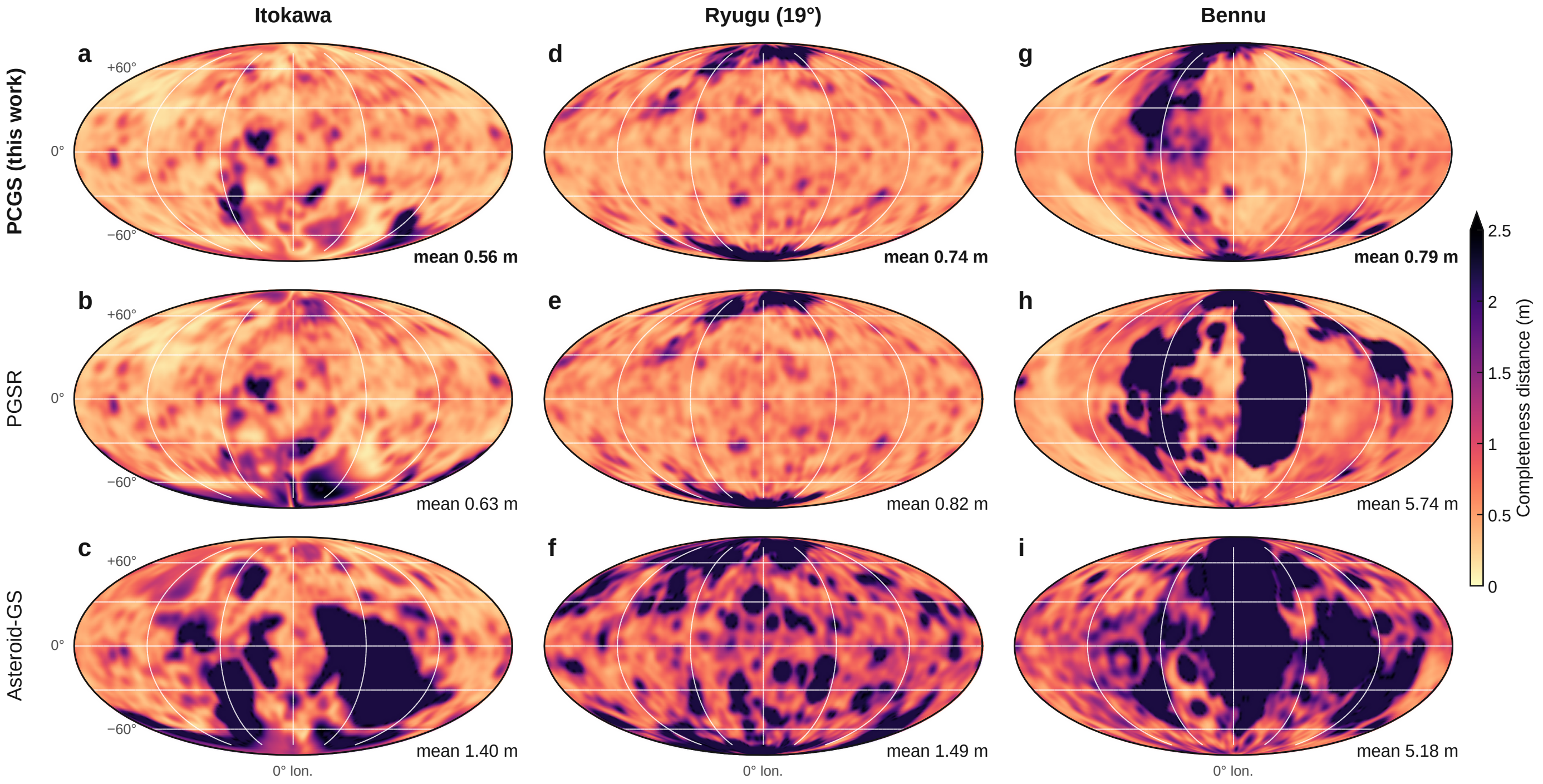}
\caption{Global completeness distance maps on Itokawa (a--c), Ryugu ($\alpha=19^\circ$; d--f), and Bennu (g--i) for
the first split in Mollweide projection ($0^\circ$ central meridian; $+60^\circ$, $0^\circ$, and $-60^\circ$
parallels marked on the $60^\circ\times 30^\circ$ graticule). Two-degree bin means are smoothed with a
$\sigma=1.2$-bin Gaussian kernel; labels state unsmoothed means. Asteroid-GS uses camera-ring extent on Bennu.}
\label{fig:maps}
\end{figure*}

First-split meshes clarify boulder-scale differences (Fig.~\ref{fig:recon}a--h; held-out Ryugu comparisons in
Appendix~\ref{app:details}, Figs.~\ref{fig:detail_ryugu19} and~\ref{fig:detail_ryugu40}). All three capture Itokawa's
two lobes and Ryugu's equatorial ridge \citep{Watanabe2019,Sugita2019}, yet close-ups (Fig.~\ref{fig:recon}b,f) show
\pcgs\ resolving boulders and crater rims cleanly, while PGSR introduces linear streaks and Asteroid-GS smooths fine
relief with 1--3~m outward offsets (Fig.~\ref{fig:recon}c,d,g,h). Mean normal error therefore favours smoother meshes
such as Asteroid-GS$^\dagger$ ($12.7^\circ$ Itokawa, $11.3^\circ$ Ryugu), RaDe-GS ($13.6^\circ$ Itokawa), and COLMAP
dense MVS ($11.9^\circ$ vs.\ $12.1^\circ$ for \pcgs\ on Ryugu), as over-smoothing damps high-frequency normal variance
despite metre-scale bias, whereas \pcgs\ retains steep boulder relief plus minor sub-shading normal noise.

Bennu provides an image-independent test against the OLA v21 laser-altimeter model across $\alpha=26^\circ$--$60^\circ$,
where up to half the disc is unlit. Here \pcgs\ separates sharply from all competing Gaussian-splatting baselines
(Table~\ref{tab:main}; Fig.~\ref{fig:numbers}c): RaDe-GS degrades to $1.21\pm 0.33$~m mean ($4.76$~m RMSE, $3.72$~m
Chamfer; non-watertight, $-23.0\%$ area deficit), PGSR and Asteroid-GS$^\dagger$ to $3.78$~m ($3.84$ and $5.94$~m
Chamfer), and 2DGS to $26.77$~m. By contrast, \pcgs\ achieves $0.50\pm 0.01$~m mean, $0.78\pm 0.02$~m RMSE,
$0.76\pm 0.01$~m Chamfer, F@1~m $0.794\pm 0.010$, and $31.6\pm 0.3^\circ$ normal error on a watertight mesh, plus the
highest test PSNR ($22.7\pm 0.6$~dB). COLMAP dense MVS is a strong geometric baseline ($0.53\pm 0.01$~m mean,
$0.71$~m RMSE, $0.67$~m Chamfer, F@1~m $0.818$, $29.1\pm 0.3^\circ$), matching \pcgs\ in mean error with slightly lower
RMSE and Chamfer, but smooths boulder relief more heavily ($-16.4\pm 0.3\%$ vs.\ $-12.3\pm 0.3\%$ area deficit), leaves
one of three splits non-watertight, and provides no appearance model or NVS. First-split meshes illustrate the
splatting failure (Fig.~\ref{fig:recon}i--l): unlit regolith blends with sky, so PGSR develops holes and outward
patches (17.3\% $>3$~m outside) and Asteroid-GS$^\dagger$ recedes inwards (11.3\% $>3$~m inside), whereas \pcgs\ --
via detached Sun-view depth and sunlit-consistent consensus -- avoids grazing-shadow spikes and stays zero-centred
($0.51$~m mean, $0.81$~m RMSE, $0.7\%$ $|d|>3$~m). The ${\sim}29^\circ$--$32^\circ$ normal error shared by COLMAP MVS
and \pcgs\ reflects steep metre-scale facets in the $0.88$~m OLA model varying below the 2--3-pixel lateral limit of
$0.62$--$2.60$~m per pixel imagery (Fig.~\ref{fig:recon}j).

\begin{figure*}[!t]
\centering
\includegraphics[width=\textwidth]{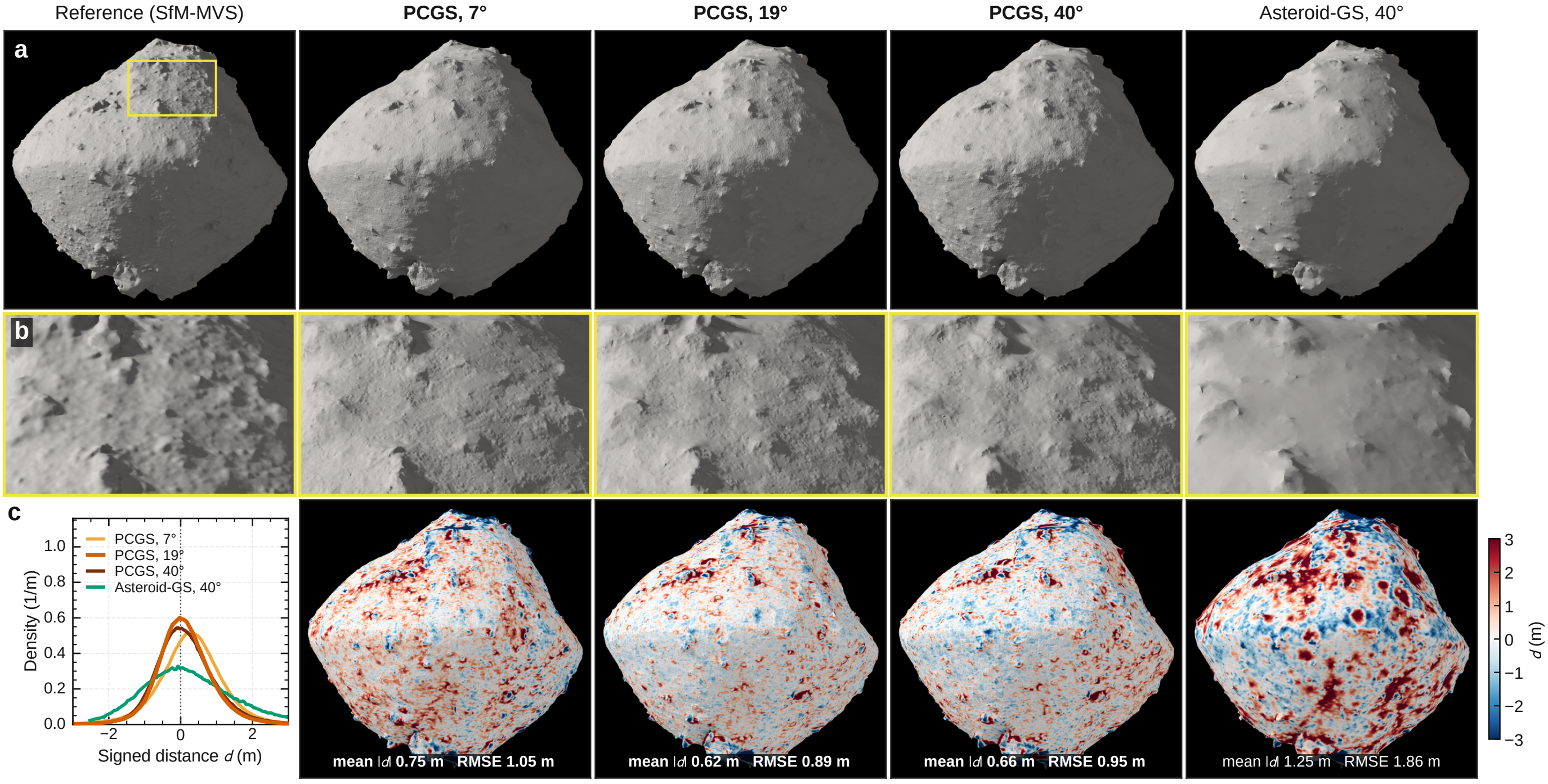}
\caption{Cross-phase consistency on Ryugu for \pcgs\ ($\alpha=7^\circ$, $19^\circ$, and $40^\circ$; first split)
and Asteroid-GS ($40^\circ$), rendered from the $19^\circ$ pose of Fig.~\ref{fig:recon}e. (a) Global shaded views.
(b) Boulder close-ups from Fig.~\ref{fig:recon}f. (c) Signed distance distributions and maps ($\pm 3$~m) with mean
$|d|$ and RMSE.}
\label{fig:ryuguphase}
\end{figure*}

Spatial completeness errors track observing and lighting geometry across all three bodies (Figs.~\ref{fig:curves}e,f
and~\ref{fig:maps}). On Itokawa, \pcgs\ stays uniform with latitude outside poorly lit neck and south-polar
concavities, whereas PGSR degrades southward and Asteroid-GS shows broad $>2.5$~m zones. On Bennu, mean completeness
distance is $0.79$~m for \pcgs\ vs.\ $5.74$~m (PGSR), $5.18$~m (Asteroid-GS$^\dagger$), and $6.27\pm 2.89$~m (RaDe-GS;
Fig.~\ref{fig:maps}g--i). On Ryugu, \pcgs\ and PGSR stay within 0.5--0.7~m at low--mid latitudes (Asteroid-GS $>1.0$~m
nearly everywhere) before all three rise to 2.5--3.7~m poleward of $\pm 60^\circ$, reflecting near-equatorial viewing
and lighting (weak meridional shading; Sects.~\ref{sec:render} and~\ref{sec:images}) plus higher polar uncertainty in
the Ryugu reference \citep{ChenH2025}.

\begin{table}[!t]
\caption{Enclosed volume errors of \pcgs\ meshes relative to the mission reference shape models.}
\label{tab:volume}
\centering
\footnotesize\setlength{\tabcolsep}{5.04pt}
\begin{tabular}{@{}lccc@{}}
\toprule
 & & \multicolumn{2}{c}{Volume error (\%)} \\
\cmidrule(lr){3-4}
Image set & Scale & Registered & Trajectory scale \\
\midrule
Ryugu, 7$^\circ$      & 0.9993 & $+0.11$ & $+0.32$ \\
Itokawa, 10$^\circ$   & 1.0105 & $-0.19$, $-0.27$, $-0.25$ & $-3.27$, $-3.35$, $-3.33$ \\
Ryugu, 19$^\circ$     & 0.9985 & $+0.01$, $+0.01$, $+0.00$ & $+0.46$, $+0.46$, $+0.45$ \\
Ryugu, 40$^\circ$     & 0.9969 & $+0.06$, $+0.05$ & $+1.00$, $+0.99$ \\
Bennu, 26--60$^\circ$ & 0.9976 & $+0.37$, $+0.38$, $+0.37$ & $+1.09$, $+1.10$, $+1.08$ \\
\bottomrule
\end{tabular}
\end{table}

Every \pcgs\ mesh is closed (zero boundary edges; $\leq 5$ non-manifold edges among ${\sim}10^7$), enabling exact
volume comparisons $(V_\mathrm{rec}-V_\mathrm{ref})/V_\mathrm{ref}$ (Table~\ref{tab:volume}). After shared SfM
registration, volumes agree within 0.01\% on Ryugu ($19^\circ$), 0.11\% on Ryugu ($7^\circ$, $40^\circ$), 0.3\% on
Itokawa, and $0.38\%$ on Bennu ($+0.37\pm 0.01\%$); at nominal SPICE scale, errors stay within $-3.4\%$ on Itokawa and
$+1.1\%$ on Ryugu and Bennu ($+0.5\%$ at $19^\circ$, $+1.0\%$ at $40^\circ$, $+1.1\%$ on Bennu). Among baselines, only
GOF consistently closes (registered errors $-1.4\%$ to $-3.0\%$ on Itokawa, $+0.5\%$ to $+0.6\%$ on Ryugu), RaDe-GS is
open on every split, and COLMAP MVS leaves openings on 5/11 splits ($0.01\%$--$0.35\%$ when closed).

\begin{table*}[!t]
\centering
\caption{Shape accuracy across all five image sets ordered by solar phase angle (first split; mean vertex distance in metres and F-score at 1~m).}
\label{tab:phase}
\footnotesize
\setlength{\tabcolsep}{4.48pt}
\begin{tabular}{lrrrrrrrrrr}
\toprule
 & \multicolumn{2}{c}{Ryugu, 7$^\circ$} & \multicolumn{2}{c}{Itokawa, 10$^\circ$} & \multicolumn{2}{c}{Ryugu, 19$^\circ$} & \multicolumn{2}{c}{Ryugu, 40$^\circ$} & \multicolumn{2}{c}{Bennu, 26--60$^\circ$} \\
\cmidrule(lr){2-3}\cmidrule(lr){4-5}\cmidrule(lr){6-7}\cmidrule(lr){8-9}\cmidrule(lr){10-11}
Method & Mean (m) & F@1~m & Mean (m) & F@1~m & Mean (m) & F@1~m & Mean (m) & F@1~m & Mean (m) & F@1~m \\
\midrule
COLMAP MVS & -- & -- & 0.85 & 0.805 & 1.12 & 0.721 & 1.16 & 0.668 & \underline{0.55} & \textbf{0.837} \\
PGSR & 0.84 & \textbf{0.730} & 0.90 & 0.812 & 0.72 & 0.789 & 0.92 & 0.711 & 4.06 & 0.642 \\
Asteroid-GS & 1.45 & 0.491 & 1.10 & 0.621 & 1.29 & 0.519 & 1.25 & 0.527 & 2.85 & 0.408 \\
PCGS without phase switch & \textbf{0.74} & \underline{0.720} & 0.68 & 0.841 & \underline{0.64} & 0.779 & \underline{0.68} & \underline{0.743} & 2.44 & 0.606 \\
PCGS without night-side band & 0.76 & 0.708 & \textbf{0.58} & \underline{0.846} & \textbf{0.62} & \underline{0.791} & 0.71 & 0.709 & 4.18 & 0.444 \\
\rowcolor{pcgsrow} \textbf{PCGS (this work)} & \underline{0.75} & 0.712 & \underline{0.60} & \textbf{0.849} & \textbf{0.62} & \textbf{0.798} & \textbf{0.66} & \textbf{0.757} & \textbf{0.51} & \underline{0.788} \\
\bottomrule
\end{tabular}
\tablefoot{Bold and underlined values mark the best and second-best entries per column. Asteroid-GS uses the shared object extent except on Bennu, where its default camera-ring extent is used. COLMAP MVS was not evaluated on Ryugu at 7$^\circ$ (marked --).}
\end{table*}

\subsection{Sensitivity to solar phase angle}
\label{sec:phase}

Across $\alpha=7^\circ$--$60^\circ$, \pcgs\ keeps first-split mean errors within $0.51$--$0.75$~m, outperforming PGSR
($0.72$--$4.06$~m) and Asteroid-GS ($1.10$--$2.85$~m) in every regime, beating COLMAP dense MVS by 30\%--45\% on
Itokawa ($0.60$ vs.\ $0.85$~m) and Ryugu at $19^\circ$ ($0.62$ vs.\ $1.12$~m) and $40^\circ$ ($0.66$ vs.\ $1.16$~m;
$1.17\pm 0.01$~m over two splits), and matching it on Bennu ($0.51$ vs.\ $0.55$~m; Table~\ref{tab:phase}). Across the
three 20~km Ryugu epochs, error is lowest at $19^\circ$ ($0.62$~m), rising modestly to $0.75$~m at $7^\circ$ (weak
shading) and $0.66$~m at $40^\circ$ (wide dark crescent), while all three models reproduce consistent global and
boulder relief (Figs.~\ref{fig:ryuguphase}, \ref{fig:detail_ryugu19}, and~\ref{fig:detail_ryugu40}) and registered
volumes within 0.10\% (Table~\ref{tab:volume}). At $40^\circ$, Asteroid-GS ($1.25$~m mean, $1.86$~m RMSE) and COLMAP
MVS ($1.17$~m mean, $1.81$~m RMSE) degrade markedly.

Although inferred from linearly scaled or archived-DN images rather than radiometrically calibrated data, the effective
lunar-Lambert weight $L$ varies systematically with phase angle: on Ryugu it drops monotonically from $0.78$ ($7^\circ$)
to $0.62$ ($19^\circ$) and $0.37$ ($40^\circ$; spread $<0.01$), with Itokawa ($10^\circ$, $L=0.64$) lying between
Ryugu's $7^\circ$ and $19^\circ$ values. This shift from near-Lommel--Seeliger scattering towards stronger limb
darkening matches the empirical lunar trend \citep{McEwen1991}; direct numerical comparison with \citet{Tatsumi2020}
and \citet{Golish2021} is precluded by their explicit phase functions. On Bennu ($\alpha=26^\circ$--$60^\circ$), $L$
converges to ${\approx}0.1$ as an effective parameter absorbing inter-frame phase darkening.

\subsection{Novel-view synthesis versus geometric fidelity}
\label{sec:nvs}

\begin{figure*}[!t]
\centering
\includegraphics[width=\textwidth]{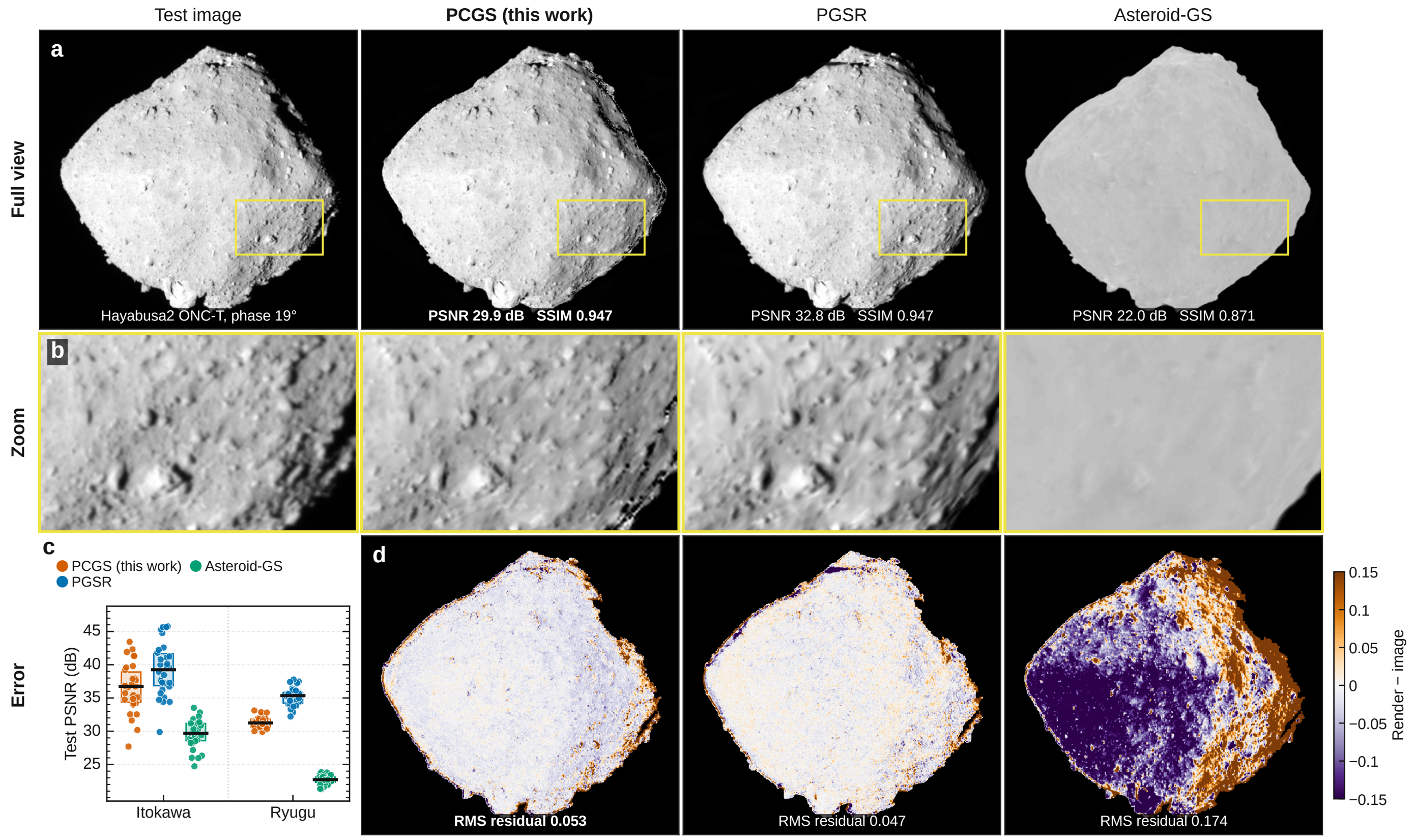}
\caption{Decoupling of novel-view synthesis from 3D shape accuracy. (a) Ryugu test view ($\alpha=19^\circ$) and
gain- and offset-matched renders (PSNR, SSIM); yellow boxes mark (b) terminator boulder close-ups. (c) Per-view
PSNR across three splits ($n=30$ views); horizontal bars and shaded boxes mark medians and interquartile ranges. (d)
Foreground residuals ($\text{render}-\text{image}$) with RMS values.}
\label{fig:nvs}
\end{figure*}

High novel-view synthesis (NVS) fidelity does not imply accurate 3D geometry (Fig.~\ref{fig:nvs};
Table~\ref{tab:images_full}; COLMAP dense MVS produces geometry only). On Ryugu ($19^\circ$), PGSR matches test
appearance closely (Fig.~\ref{fig:nvs}a,b), \pcgs\ preserves boulder relief and diffuse shading without cast shadows
($V_p=1$ on unseen views; Sect.~\ref{sec:render}), and Asteroid-GS blurs fine detail without fitted test codes. Among
surface-oriented splatting baselines, RaDe-GS attains top PSNR on Itokawa ($40.0$~dB) and Ryugu ($35.8$~dB; PGSR
$39.3$/$35.2$~dB, \pcgs\ $36.5$/$31.3$~dB), yet \pcgs\ leads SSIM on Ryugu ($0.952$), ranks second in LPIPS on Itokawa
($0.098$ vs.\ $0.096$) and tied second on Ryugu ($0.225$), and leads test PSNR on Bennu ($22.7$ vs.\ $20.7$~dB RaDe-GS
and $20.3$~dB PGSR). \pcgs's lower Itokawa/Ryugu PSNR reflects both removing free view-dependent colour and unshadowed
test renders ($V_p=1$), whose positive shadow residuals (Fig.~\ref{fig:nvs}d) make test PSNR a conservative lower
bound. Decisively, top NVS scores coexist with geometric collapse: RaDe-GS leads Ryugu PSNR ($35.8$~dB, SSIM $0.950$)
while collapsing to $3.12\pm 0.85$~m mean ($3.43$~m Chamfer, F@1~m $0.249$), GOF scores $38.0$~dB on Itokawa at
$2.40$~m mean, and Mip-Splatting reaches $42.6$~dB at $2.48$~m mean and $73.26$~m Chamfer
(Tables~\ref{tab:shape_full} and~\ref{tab:images_full}), showing NVS metrics cannot proxy shape accuracy.

\subsection{Ablation study}
\label{sec:ablation}

\begin{table}[!t]
\centering
\caption{Leave-one-out ablation study on the first split.}
\label{tab:ablation}
\scriptsize
\renewcommand{\arraystretch}{1.05}
\setlength{\tabcolsep}{3.16pt}
\begin{tabular}{lrrrrr}
\toprule
Variant & Mean (m) & RMSE (m) & F@1~m & Normal ($^\circ$) & PSNR (dB) \\
\midrule
\multicolumn{6}{l}{\textit{Itokawa, 10$^\circ$}} \\
\rowcolor{pcgsrow} \textbf{PCGS (full model)} & 0.60 & 1.07 & 0.849 & 14.8 & 37.2 \\
Unshaded albedo & 0.89 & 2.46 & 0.799 & 16.3 & 39.1 \\
View-dependent colour & 0.78 & 2.21 & 0.820 & 15.6 & 41.5 \\
No phase switch & 0.68 & 2.47 & 0.841 & 14.4 & 38.8 \\
No night-side band & 0.58 & 0.92 & 0.846 & 14.4 & 36.8 \\
No silhouette loss & 0.97 & 2.29 & 0.762 & 16.5 & 32.5 \\
No multi-view NCC & 1.26 & 3.07 & 0.627 & 20.4 & 37.5 \\
No pixel-conv.\ correction & 0.63 & 1.27 & 0.849 & 15.8 & 37.1 \\
\addlinespace[1.5pt]
\multicolumn{6}{l}{\textit{Ryugu, 19$^\circ$}} \\
\rowcolor{pcgsrow} \textbf{PCGS (full model)} & 0.62 & 0.89 & 0.798 & 12.1 & 31.4 \\
Unshaded albedo & 0.70 & 1.00 & 0.745 & 13.1 & 33.8 \\
View-dependent colour & 0.65 & 1.02 & 0.785 & 12.8 & 35.5 \\
No phase switch & 0.64 & 0.93 & 0.779 & 12.6 & 33.4 \\
No night-side band & 0.62 & 0.90 & 0.791 & 11.9 & 30.0 \\
No silhouette loss & 0.64 & 0.98 & 0.791 & 12.5 & 30.3 \\
No multi-view NCC & 0.82 & 1.21 & 0.694 & 13.7 & 30.6 \\
No pixel-conv.\ correction & 0.65 & 0.93 & 0.785 & 15.0 & 31.0 \\
\addlinespace[1.5pt]
\multicolumn{6}{l}{\textit{Ryugu, 40$^\circ$}} \\
\rowcolor{pcgsrow} \textbf{PCGS (full model)} & 0.66 & 0.95 & 0.757 & 12.8 & 33.0 \\
Unshaded albedo & 0.92 & 2.33 & 0.664 & 14.1 & 33.2 \\
View-dependent colour & 0.70 & 1.14 & 0.739 & 13.1 & 36.9 \\
No phase switch & 0.68 & 1.02 & 0.743 & 13.1 & 35.5 \\
No night-side band & 0.71 & 1.07 & 0.709 & 12.5 & 28.1 \\
No silhouette loss & 5.39 & 39.46 & 0.738 & 13.0 & 28.0 \\
No multi-view NCC & 0.96 & 1.50 & 0.623 & 14.7 & 32.1 \\
\addlinespace[1.5pt]
\multicolumn{6}{l}{\textit{Bennu, 26--60$^\circ$}} \\
\rowcolor{pcgsrow} \textbf{PCGS (full model)} & 0.51 & 0.81 & 0.788 & 31.9 & 23.4 \\
View-dependent colour & 4.72 & 12.63 & 0.496 & 37.2 & 20.7 \\
No phase switch & 2.44 & 7.63 & 0.606 & 36.6 & 22.8 \\
No night-side band & 4.18 & 10.28 & 0.444 & 34.8 & 19.8 \\
\bottomrule
\end{tabular}
\end{table}

Table~\ref{tab:ablation} reports leave-one-out ablations on the first split of Itokawa ($10^\circ$), Ryugu
($19^\circ$, $40^\circ$), and Bennu ($26^\circ$--$60^\circ$); Appendix~\ref{app:tables} adds pre-smoothing extraction
stages (Table~\ref{tab:pipeline}) and phase-switch tests without terminator masking (Table~\ref{tab:priors}).

\subsubsection{Photometric rendering}
\label{sec:abl_photo}

Removing physical shading degrades geometry across every target (Table~\ref{tab:ablation}). Unshaded albedo (`Unshaded
albedo') raises mean distance by 48\% on Itokawa ($0.60\to 0.89$~m; RMSE $1.07\to 2.46$~m), 14\% on Ryugu ($19^\circ$,
to $0.70$~m), and 38\% at $40^\circ$ ($0.92$~m). Reverting to 2DGS spherical harmonics (`View-dependent colour') raises
mean error by 30\% on Itokawa ($0.78$~m; RMSE $2.21$~m) and 6\% on Ryugu ($0.65$ and $0.70$~m at $19^\circ$, $40^\circ$)
while inflating test PSNR by 3.9--4.3~dB, confirming appearance overfitting at the cost of shape; on Bennu it fails
altogether even with the night-side band ($4.72$~m mean, $12.63$~m RMSE, worse than PGSR at $4.06$~m).

Because both variants also disable the phase switch, comparing them with `No phase switch' isolates lunar-Lambert
shading alone: on Itokawa it cuts mean error from $0.89$ (unshaded) and $0.78$~m (SH) to $0.68$~m, normal error from
$16.3^\circ$ to $14.4^\circ$, and raises F@1~m from $0.799$ to $0.841$ at nearly constant RMSE ($2.46$ vs.\ $2.47$~m),
showing shading governs bulk slopes while TV suppresses outliers. On Ryugu ($19^\circ$), shading lowers SH from $0.65$
to $0.64$~m. Even with SH, the remaining \pcgs\ pipeline beats PGSR on split~1 ($0.78$ vs.\ $0.90$~m Itokawa; $0.65$
vs.\ $0.72$~m Ryugu), attributing 60\% of our Itokawa gain over PGSR ($0.18/0.30$~m) and 30\% on Ryugu ($0.03/0.10$~m)
to photometry and the phase switch.

\subsubsection{Multi-view, silhouette, and terminator constraints}

\begin{figure*}[!t]
\centering
\includegraphics[width=\textwidth]{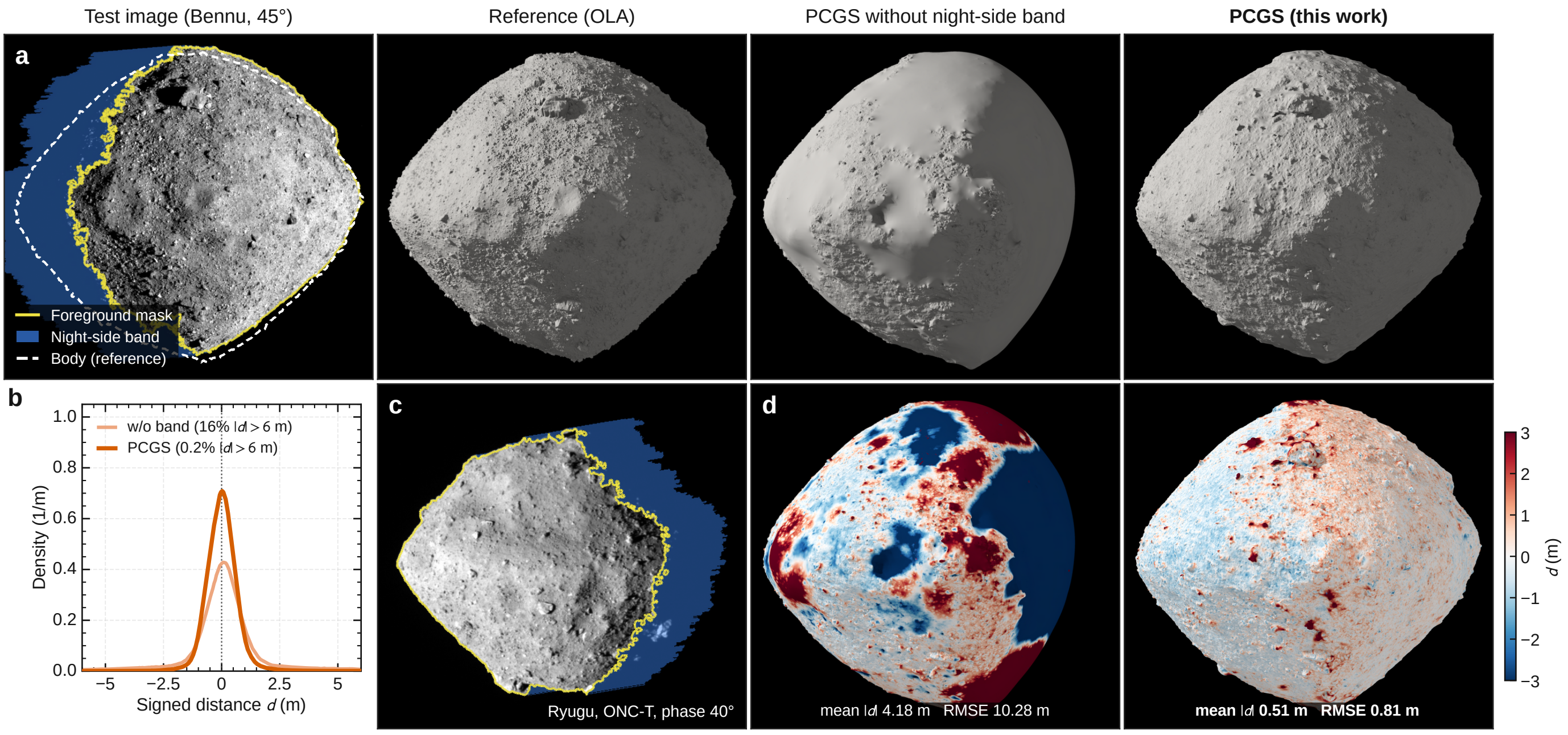}
\caption{Night-side band at high phase angles (first split). (a) Bennu test image ($\alpha=45^\circ$) with sunlit
mask $M$ (yellow), excluded band (blue; Eq.~\eqref{eq:night}), and OLA limb (dashed white), alongside the OLA reference
and \pcgs\ without and with the band. (b) Bennu signed distance distributions annotated with $|d|>6$~m fractions. (c)
Mask and band on Ryugu ($\alpha=40^\circ$). (d) Signed distance maps on Bennu ($\pm 3$~m).}
\label{fig:night}
\end{figure*}

At low to moderate phase, removing multi-view NCC causes the sharpest single-term drop, raising mean error to $1.26$~m
on Itokawa (F@1~m $0.849\to 0.627$), $0.82$~m on Ryugu ($19^\circ$; $0.798\to 0.694$), and $0.96$~m at $40^\circ$.
Removing the silhouette loss degrades Itokawa ($0.97$~m) and Ryugu ($19^\circ$, $0.64$~m) and leaves unpruned
background surfels at $40^\circ$ (RMSE $0.95\to 39.5$~m).

The night-side band governs high-phase convergence (Fig.~\ref{fig:night}). At $\alpha=10^\circ$ (Itokawa), the narrow
band slightly relaxes terminator supervision (omitting it shifts split-1 mean $0.60\to 0.58$~m, RMSE $1.07\to 0.92$~m).
At high phase, enabling the band cuts mean error from $0.71$ to $0.66$~m on Ryugu ($40^\circ$; F@1~m $0.709\to 0.757$;
Fig.~\ref{fig:night}c) and eightfold on Bennu, from $4.18$ to $0.51$~m (without it, \pcgs\ trails PGSR at $4.06$~m and
Asteroid-GS$^\dagger$ at $2.85$~m; Table~\ref{tab:phase}), preventing dark-side carving and lit-side bulging (16\% vs.\
$0.2\%$ of vertices at $|d|>6$~m; Fig.~\ref{fig:night}a,b,d).

\subsubsection{Phase switch and mesh extraction}
\label{sec:rule}

\begin{figure*}[!t]
\centering
\includegraphics[width=\textwidth]{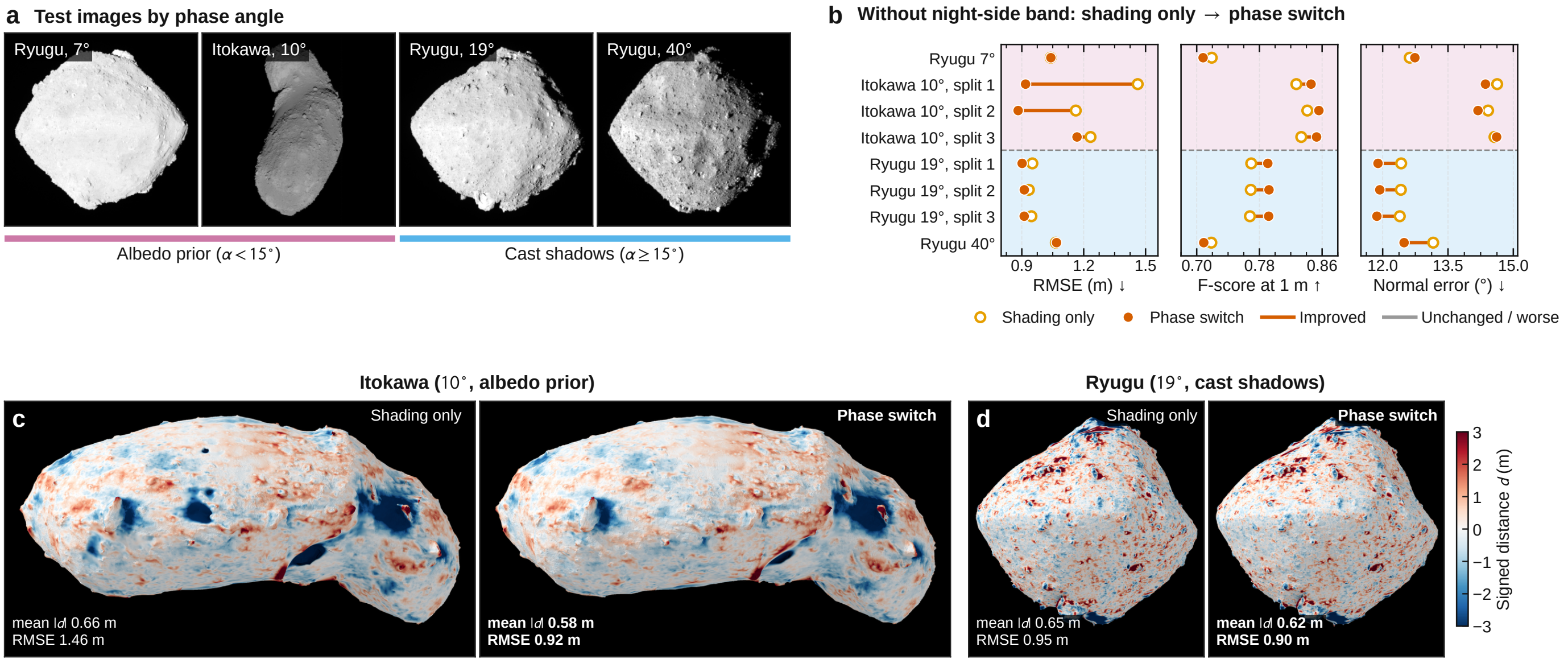}
\caption{Effect of the phase switch without the night-side band. (a) Test images ordered by solar phase angle with
active branch bars ($\alpha<15^\circ$: albedo TV prior; $\alpha\geq 15^\circ$: cast shadows). (b) Split-by-split
shifts from shading alone (open circles) to shading with the phase switch (filled circles; vermilion: improvement,
grey: deterioration). (c, d) First-split signed distance maps ($\pm 3$~m) on Itokawa and Ryugu ($19^\circ$).}
\label{fig:priors}
\end{figure*}

With the night-side band active (Tables~\ref{tab:phase} and~\ref{tab:ablation}), the phase switch helps on four of five
campaigns: removing it raises mean error to $0.68$~m (Itokawa), $0.64$ and $0.68$~m (Ryugu $19^\circ$, $40^\circ$), and
$2.44$~m (Bennu), with a $0.01$~m opposite shift at $7^\circ$. Testing branches without the band clarifies the
mechanism (Fig.~\ref{fig:priors}; Table~\ref{tab:priors}): albedo TV cuts Itokawa RMSE across all splits
($1.46\to 0.92$~m on split~1) by keeping weak shading out of albedo, while cast shadows improve Ryugu ($19^\circ$) on
every split ($0.65\to 0.62$~m) and cut Bennu RMSE by 63\% ($27.49\to 10.28$~m). Without the band at $7^\circ$ and
$40^\circ$, shifts stay within $0.02$~m ($40^\circ$ shadows improve normals $0.7^\circ$, F@1~m $0.719\to 0.709$),
whereas enabling both priors degrades shadow-rich frames as TV damps shadow contrast; the data constrain the switch
threshold to $10.4^\circ$--$18.5^\circ$.

Omitting COLMAP pixel-centre alignment (`No pixel-convention correction'; Table~\ref{tab:ablation}) adds only $0.03$~m
to mean error on Itokawa ($0.63$~m) and Ryugu ($19^\circ$, $0.65$~m, still below PGSR's $0.72$~m) after tie-point
alignment, but degrades Ryugu normals ($12.1^\circ\to 15.0^\circ$). Stepping through pre-smoothing extraction
(Table~\ref{tab:pipeline}) shows pixel alignment cuts raw Ryugu mean error from $1.35$ to $0.76$~m, biharmonic fairing
cuts Itokawa RMSE from $1.71$ to $1.29$~m across concavities, and the tie-point normal offset removes the $0.1$--$0.2$-pixel camera-ward shift ($0.76\to 0.66$~m on Ryugu).

\Needspace{8\baselineskip}
\section{Discussion}
\label{sec:discussion}

\subsection{Accuracy relative to the image resolution}

Across all five datasets, \pcgs\ attains mean errors at or below one image pixel: $0.59$~m on Itokawa ($0.78$~pixels at
$0.76$~m~px$^{-1}$), $0.62$--$0.75$~m on Ryugu ($0.28$--$0.34$~pixels at $2.15$--$2.36$~m~px$^{-1}$; $0.29$~pixels at
$19^\circ$), and $0.50$~m on Bennu ($0.19$--$0.81$~pixels across $0.62$--$2.60$~m~px$^{-1}$). Triangulation and
multi-azimuth shading constrain elevation below the ground sample distance, whereas lateral resolution stays near
2--3~pixels \citep{Palmer2022}. This lateral limit smooths steep metre-scale boulders resolved in the $0.88$~m OLA
model (Fig.~\ref{fig:recon}j), explaining why every optical reconstruction of Bennu -- including COLMAP dense MVS
($29.1^\circ$) and \pcgs\ ($31.6^\circ$) -- shows ${\sim}30^\circ$ facet-level normal scatter against OLA despite
sub-metre positional accuracy. On Itokawa, $0.59$~m matches the $0.60$~m mean facet spacing of the \citet{Gaskell2008}
SPC reference.

\subsection{Comparison with state-of-the-art neural models}
\label{sec:published}

From 50/10 splits of the same Itokawa and Ryugu ($19^\circ$) campaigns, \citet{Zhang2025AsteroidGS} reported
Asteroid-GS mean errors of $0.705$~m and $1.341$~m under their own protocol, vs.\ $0.94$~m and $1.29$~m for
Asteroid-GS$^\dagger$ here (Table~\ref{tab:main}); Ryugu agrees closely, while the Itokawa offset reflects unpublished
splits/evaluation code and dilated-mask pruning. Separately, using 70 Box-A (${\sim}2.2$~m~px$^{-1}$) or 61 Box-C
($0.6$--$0.7$~m~px$^{-1}$) frames plus explicit SfM-MVS 3D supervision ($\leq 1$~pixel residual), neural implicit SDF
modelling of Ryugu \citep{ChenH2025} placed $80.7\%$ and $95.5\%$ of vertices within 1 and 2~m. Without dense 3D
supervision, \pcgs\ places $82.5\%$ and $97.1\%$ within 1 and 2~m on Ryugu ($19^\circ$) from 50 Box-A frames in ${\sim}30$~min
rather than multi-hour training; because splits and protocols differ, these indicate comparable accuracy rather than a
strictly controlled ranking.

\subsection{Role of solar illumination}

Photometric shading contributes most where narrow-baseline stereo matching is weakest -- on smooth, low-texture
regolith. Reverting to spherical harmonics degrades mean error far more on Itokawa ($+30\%$) than on boulder-rich Ryugu
($+6\%$; Sect.~\ref{sec:abl_photo}), though pixel-scale and reference differences preclude attributing this contrast
solely to roughness. At Bennu's high phase angles ($26^\circ$--$60^\circ$), where cast shadows and wide dark crescents
curtail stereo overlap, unconstrained view-dependent colour fails altogether ($4.72$~m mean). Because full-system
benchmarks couple photometry with silhouette masking and extraction choices (Sect.~\ref{sec:implementation}),
controlled ablations are essential to isolate the photometric gain. At the finest scales, narrow dark albedo halos
beside Ryugu boulders (Fig.~\ref{fig:overview}c) show that sub-boulder shadows and micro-roughness still leak partially
into scalar albedo $A$, confining reliable photometric slope control to multi-pixel scales.

\subsection{Limitations and future work}

Two primary limitations merit note. First, the photometric and geometric formulation retains simplifying assumptions:
Eq.~\eqref{eq:ll} uses a single global parameter $L$ without an explicit phase function or per-frame exposure terms,
relies on the heuristic $15^\circ$ phase switch (leaving held-out Ryugu at $7^\circ$ at $0.75$~m, \pcgs's largest mean
error; Table~\ref{tab:phase}), slightly damps bright albedo features (such as Itokawa's Muses-C Regio) via TV
regularisation, and uses fixed metric tolerances (Table~\ref{tab:settings}, spanning $0.2$--$5$~pixels here) that
should scale with pixel size for much larger or smaller bodies. Second, operational viewing, scale, and benchmarking
constraints bound current evaluation: without external registration, nominal SPICE-scale volume errors ($-3.35\%$ to
$+1.10\%$; Table~\ref{tab:volume}) propagate into early bulk-density estimates; unviewed polar caps (needing
higher-latitude views) and faired concavities must be flagged as unmeasured, while the optical Itokawa and Ryugu
references carry higher polar uncertainty than Bennu's laser-altimeter OLA model \citep{ChenH2025}; and all five test
sets provide full-rotation SPICE-initialised coverage rather than partial-rotation flybys, with competing neural
baselines run as released and multi-hour neural SDFs \citep{ChenH2024,ChenS2024,ChenH2025} not re-benchmarked on
identical splits.

Future work should replace the heuristic $15^\circ$ switch with a continuous multi-phase law such as Hapke's
\citep{Hapke2012} -- preventing phase-darkening bias in $L$ -- make geometric tolerances pixel-scale adaptive, and test
partial-rotation flyby arcs and larger initial pointing errors. From Box-A imagery, SPICE ephemerides, and COLMAP poses
alone, \pcgs\ already builds closed global models in ${\sim}30$~min on one GPU, aiding early density, gravity, and
hazard assessments for ESA's Hera \citep{Michel2022}.

\FloatBarrier
\Needspace{10\baselineskip}
\section{Conclusions}
\label{sec:conclusions}

We have presented \pcgs, a photometrically constrained two-dimensional Gaussian splatting method that builds closed
global asteroid shape models from 40--50 full-rotation early-rendezvous images in roughly 30~min on a single GPU by
combining lunar-Lambert shading of view-independent relative scalar albedos under SPICE solar vectors, terminator
night-side protection, a $15^\circ$ albedo-TV/detached cast-shadow phase switch, and depth-consensus meshing. Across
five campaigns of Itokawa, Ryugu, and Bennu ($\alpha=7^\circ$--$60^\circ$), \pcgs\ attains sub-metre mean errors
($0.50$--$0.75$~m) and enclosed volumes within $0.38\%$ registered ($3.4\%$ at nominal SPICE scale), outperforming all
five Gaussian-splatting baselines and improving on classical COLMAP dense MVS in Itokawa/Ryugu accuracy (31\%--45\%),
Bennu boulder surface-area retention ($-12.3\%$ vs.\ $-16.4\%$), watertight closure, and joint novel-view synthesis.
Controlled ablations confirm that free view-dependent colour overfits brightness at the expense of geometry whereas
terminator protection prevents high-phase collapse; with unviewed polar caps and faired concavities flagged as
unmeasured, \pcgs\ offers a practical rapid-modelling complement to late-mission SPG, SPC, and altimetry, while
continuous multi-phase reflectance and partial-rotation flyby tests remain for future work on missions such as ESA's
Hera \citep{Michel2022}.

\section*{Data availability}
The \pcgs\ code and evaluation scripts are publicly available on GitHub
(\url{https://github.com/Xin-YuXiao/AsteroidModeling}); frame lists, splits, refined poses, masks, and reconstructed
meshes will be archived on Zenodo upon publication. Spacecraft images, SPICE kernels, and reference models are publicly
available from JAXA's DARTS and the NASA PDS Small Bodies Node.

\begin{acknowledgements}
This work was supported by the National Natural Science Foundation of China (Grant No.~42388101) and used archival data
and reference models from JAXA DARTS, NASA PDS Small Bodies Node, and NASA NAIF.
\end{acknowledgements}

\clearpage
\bibliographystyle{aa}
\bibliography{refs}

\begin{thebibliography}{67}
\expandafter\ifx\csname natexlab\endcsname\relax\def\natexlab#1{#1}\fi

\bibitem[{{Acton}(1996)}]{Acton1996}
{Acton}, C.~H. 1996, Planet. Space Sci., 44, 65

\bibitem[{{Al Asad} {et~al.}(2021){Al Asad}, {Philpott}, {Johnson},
  {et~al.}}]{AlAsad2021}
{Al Asad}, M.~M., {Philpott}, L.~C., {Johnson}, C.~L., {et~al.} 2021, PSJ, 2,
  82

\bibitem[{Alexandrov \& Beyer(2018)}]{Alexandrov2018}
Alexandrov, O. \& Beyer, R.~A. 2018, Earth Space Sci., 5, 652

\bibitem[{{Barnouin} {et~al.}(2019){Barnouin}, {Daly}, {Palmer},
  {et~al.}}]{Barnouin2019}
{Barnouin}, O.~S., {Daly}, M.~G., {Palmer}, E.~E., {et~al.} 2019, Nat. Geosci.,
  12, 247

\bibitem[{{Botsch} \& {Kobbelt}(2004)}]{Botsch2004}
{Botsch}, M. \& {Kobbelt}, L. 2004, ACM Trans. Graph., 23, 630

\bibitem[{{Chen} {et~al.}(2025{\natexlab{a}}){Chen}, {Li}, {Ye}, {Wang}, {Xie},
  {Zhai}, {Wang}, {Liu}, {Bao}, \& {Zhang}}]{Chen2024PGSR}
{Chen}, D., {Li}, H., {Ye}, W., {et~al.} 2025{\natexlab{a}}, IEEE Trans. Vis.
  Comput. Graph., 31, 6100

\bibitem[{{Chen} {et~al.}(2024{\natexlab{a}}){Chen}, {Hu}, {Willner}, {Ye},
  {Damme}, {Gl{\"a}ser}, {Zheng}, {Tong}, {Hu{\ss}mann}, \&
  {Oberst}}]{ChenH2024}
{Chen}, H., {Hu}, X., {Willner}, K., {et~al.} 2024{\natexlab{a}}, ISPRS J.
  Photogramm. Remote Sens., 212, 122

\bibitem[{{Chen} {et~al.}(2025{\natexlab{b}}){Chen}, {Willner}, {Ziese}, {Hu},
  {Huang}, {Xiao}, {Gl{\"a}ser}, {Neumann}, \& {Oberst}}]{ChenH2025}
{Chen}, H., {Willner}, K., {Ziese}, R., {et~al.} 2025{\natexlab{b}}, A\&A, 696,
  A212

\bibitem[{{Chen} {et~al.}(2024{\natexlab{b}}){Chen}, {Wu}, {Li}, {Li}, \&
  {Liu}}]{ChenS2024}
{Chen}, S., {Wu}, B., {Li}, H., {Li}, Z., \& {Liu}, Y. 2024{\natexlab{b}},
  A\&A, 687, A278

\bibitem[{{Curless} \& {Levoy}(1996)}]{Curless1996}
{Curless}, B. \& {Levoy}, M. 1996, in Proc. SIGGRAPH 96, 303

\bibitem[{Dai {et~al.}(2024)Dai, Xu, Xie, Liu, Wang, \& Xu}]{Dai2024Surfels}
Dai, P., Xu, J., Xie, W., {et~al.} 2024, in ACM SIGGRAPH 2024 Conference
  Papers, 1--11

\bibitem[{Daly {et~al.}(2017)Daly, Barnouin, Dickinson, Seabrook, Johnson,
  Cunningham, Haltigin, Gaudreau, Brunet, Aslam, Taylor, Bierhaus, Boynton,
  Nolan, \& Lauretta}]{Daly2017}
Daly, M.~G., Barnouin, O.~S., Dickinson, C., {et~al.} 2017, Space Sci. Rev.,
  212, 899

\bibitem[{{Derksen} \& {Izzo}(2021)}]{Derksen2021}
{Derksen}, D. \& {Izzo}, D. 2021, in Proc. IEEE/CVF Conference on Computer
  Vision and Pattern Recognition Workshops (CVPRW), 1152--1161

\bibitem[{{Fujiwara} {et~al.}(2006){Fujiwara}, {Kawaguchi}, {Yeomans},
  {et~al.}}]{Fujiwara2006}
{Fujiwara}, A., {Kawaguchi}, J., {Yeomans}, D.~K., {et~al.} 2006, Science, 312,
  1330

\bibitem[{Furukawa \& Ponce(2010)}]{Furukawa2010}
Furukawa, Y. \& Ponce, J. 2010, IEEE Trans. Pattern Anal. Mach. Intell., 32,
  1362

\bibitem[{Gao {et~al.}(2024)Gao, Gu, Lin, Li, Zhu, Cao, Zhang, \&
  Yao}]{Gao2024Relightable}
Gao, J., Gu, C., Lin, Y., {et~al.} 2024, in Proc. European Conference on
  Computer Vision (ECCV), 73--89

\bibitem[{{Gaskell} {et~al.}(2008){Gaskell}, {Barnouin-Jha}, {Scheeres},
  {et~al.}}]{Gaskell2008}
{Gaskell}, R.~W., {Barnouin-Jha}, O.~S., {Scheeres}, D.~J., {et~al.} 2008,
  Meteorit. Planet. Sci., 43, 1049

\bibitem[{Golish {et~al.}(2021)Golish, DellaGiustina, Li, Clark, Zou, Smith,
  Rizos, Hasselmann, Bennett, Fornasier, Ballouz, Drouet~d'Aubigny, Rizk, Daly,
  Barnouin, Philpott, Al~Asad, Seabrook, Johnson, \& Lauretta}]{Golish2021}
Golish, D.~R., DellaGiustina, D.~N., Li, J.-Y., {et~al.} 2021, Icarus, 357,
  113724

\bibitem[{Grumpe {et~al.}(2014)Grumpe, Belkhir, \& W{\"o}hler}]{Grumpe2014}
Grumpe, A., Belkhir, F., \& W{\"o}hler, C. 2014, Adv. Space Res., 53, 1735

\bibitem[{{Gu{\'e}don} \& {Lepetit}(2024)}]{Guedon2024}
{Gu{\'e}don}, A. \& {Lepetit}, V. 2024, in Proc. IEEE/CVF Conference on
  Computer Vision and Pattern Recognition (CVPR), 5354--5363

\bibitem[{{Hapke}(2012)}]{Hapke2012}
{Hapke}, B. 2012, Theory of Reflectance and Emittance Spectroscopy, 2nd edn.
  (Cambridge: Cambridge University Press)

\bibitem[{{Horn}(1990)}]{Horn1990}
{Horn}, B.~K.~P. 1990, Int. J. Comput. Vis., 5, 37

\bibitem[{{Huang} {et~al.}(2024){Huang}, {Yu}, {Chen}, {Geiger}, \&
  {Gao}}]{Huang2024}
{Huang}, B., {Yu}, Z., {Chen}, A., {Geiger}, A., \& {Gao}, S. 2024, in ACM
  SIGGRAPH 2024 Conference Papers, 32

\bibitem[{{Ishiguro} {et~al.}(2010){Ishiguro}, {Nakamura}, {Tholen},
  {et~al.}}]{Ishiguro2010}
{Ishiguro}, M., {Nakamura}, R., {Tholen}, D.~J., {et~al.} 2010, Icarus, 207,
  714

\bibitem[{Jaumann {et~al.}(2012)Jaumann, Williams, Buczkowski, Yingst,
  Preusker, Hiesinger, Schmedemann, Kneissl, Vincent, Blewett, Buratti,
  Carsenty, Denevi, De~Sanctis, Garry, Keller, Kersten, Krohn, Li, Marchi,
  Matz, McCord, McSween, Mest, Mittlefehldt, Mottola, Nathues, Neukum, O'Brien,
  Pieters, Prettyman, Raymond, Roatsch, Russell, Schenk, Schmidt, Scholten,
  Stephan, Sykes, Tricarico, Wagner, Zuber, \& Sierks}]{Jaumann2012}
Jaumann, R., Williams, D.~A., Buczkowski, D.~L., {et~al.} 2012, Science, 336,
  687

\bibitem[{Jiang {et~al.}(2024)Jiang, Tu, Liu, Gao, Long, Wang, \&
  Ma}]{Jiang2024GaussianShader}
Jiang, Y., Tu, J., Liu, Y., {et~al.} 2024, in Proc. IEEE/CVF Conference on
  Computer Vision and Pattern Recognition (CVPR), 5322--5332

\bibitem[{{Jorda} {et~al.}(2016){Jorda}, {Gaskell}, {Capanna},
  {et~al.}}]{Jorda2016}
{Jorda}, L., {Gaskell}, R., {Capanna}, C., {et~al.} 2016, Icarus, 277, 257

\bibitem[{Kaasalainen \& Torppa(2001)}]{Kaasalainen2001}
Kaasalainen, M. \& Torppa, J. 2001, Icarus, 153, 24

\bibitem[{{Kameda} {et~al.}(2017){Kameda}, {Suzuki}, {Takamatsu},
  {et~al.}}]{Kameda2017}
{Kameda}, S., {Suzuki}, H., {Takamatsu}, T., {et~al.} 2017, Space Sci. Rev.,
  208, 17

\bibitem[{{Kazhdan} \& {Hoppe}(2013)}]{Kazhdan2013}
{Kazhdan}, M. \& {Hoppe}, H. 2013, ACM Trans. Graph., 32, 29

\bibitem[{{Kerbl} {et~al.}(2023){Kerbl}, {Kopanas}, {Leimk{\"u}hler}, \&
  {Drettakis}}]{Kerbl2023}
{Kerbl}, B., {Kopanas}, G., {Leimk{\"u}hler}, T., \& {Drettakis}, G. 2023, ACM
  Trans. Graph., 42, 139

\bibitem[{Kirk {et~al.}(2003)Kirk, Howington-Kraus, Redding, Galuszka, Hare,
  Archinal, Soderblom, \& Barrett}]{Kirk2003}
Kirk, R.~L., Howington-Kraus, E., Redding, B., {et~al.} 2003, J. Geophys. Res.
  (Planets), 108, 8088

\bibitem[{{Lauretta} {et~al.}(2019){Lauretta}, {DellaGiustina}, {Bennett},
  {et~al.}}]{Lauretta2019}
{Lauretta}, D.~S., {DellaGiustina}, D.~N., {Bennett}, C.~A., {et~al.} 2019,
  Nature, 568, 55

\bibitem[{Li {et~al.}(2015)Li, Helfenstein, Buratti, Takir, \& Clark}]{Li2015}
Li, J.-Y., Helfenstein, P., Buratti, B., Takir, D., \& Clark, B.~E. 2015, in
  Asteroids IV, ed. P.~Michel, F.~E. DeMeo, \& W.~F. Bottke (Tucson: University
  of Arizona Press), 129--150

\bibitem[{Li {et~al.}(2023)Li, M{\"u}ller, Evans, Taylor, Unberath, Liu, \&
  Lin}]{Li2023Neuralangelo}
Li, Z., M{\"u}ller, T., Evans, A., {et~al.} 2023, in Proc. IEEE/CVF Conference
  on Computer Vision and Pattern Recognition (CVPR), 8456--8465

\bibitem[{Liang {et~al.}(2024)Liang, Zhang, Feng, Shan, \& Jia}]{Liang2024GSIR}
Liang, Z., Zhang, Q., Feng, Y., Shan, Y., \& Jia, K. 2024, in Proc. IEEE/CVF
  Conference on Computer Vision and Pattern Recognition (CVPR), 21644--21653

\bibitem[{{Mar{\'\i}} {et~al.}(2022){Mar{\'\i}}, {Facciolo}, \&
  {Ehret}}]{Mari2022}
{Mar{\'\i}}, R., {Facciolo}, G., \& {Ehret}, T. 2022, in Proc. IEEE/CVF
  Conference on Computer Vision and Pattern Recognition Workshops (CVPRW),
  1310--1320

\bibitem[{{Mar{\'\i}} {et~al.}(2023){Mar{\'\i}}, {Facciolo}, \&
  {Ehret}}]{Mari2023}
{Mar{\'\i}}, R., {Facciolo}, G., \& {Ehret}, T. 2023, in Proc. IEEE/CVF
  Conference on Computer Vision and Pattern Recognition Workshops (CVPRW),
  2035--2045

\bibitem[{{Martin-Brualla} {et~al.}(2021){Martin-Brualla}, {Radwan}, {Sajjadi},
  {Barron}, {Dosovitskiy}, \& {Duckworth}}]{MartinBrualla2021}
{Martin-Brualla}, R., {Radwan}, N., {Sajjadi}, M.~S.~M., {et~al.} 2021, in
  Proc. IEEE/CVF Conference on Computer Vision and Pattern Recognition (CVPR),
  7206--7215

\bibitem[{{McEwen}(1991)}]{McEwen1991}
{McEwen}, A.~S. 1991, Icarus, 92, 298

\bibitem[{{Michel} {et~al.}(2022){Michel}, {K{\"u}ppers}, {Bagatin},
  {et~al.}}]{Michel2022}
{Michel}, P., {K{\"u}ppers}, M., {Bagatin}, A.~C., {et~al.} 2022, PSJ, 3, 160

\bibitem[{{Mildenhall} {et~al.}(2020){Mildenhall}, {Srinivasan}, {Tancik},
  {Barron}, {Ramamoorthi}, \& {Ng}}]{Mildenhall2020}
{Mildenhall}, B., {Srinivasan}, P.~P., {Tancik}, M., {et~al.} 2020, in Proc.
  European Conference on Computer Vision (ECCV), 405--421

\bibitem[{{Nolan} {et~al.}(2026){Nolan}, {Driver}, \& {Christian}}]{Nolan2026}
{Nolan}, J., {Driver}, T., \& {Christian}, J. 2026, arXiv e-prints,
  arXiv:2603.11969

\bibitem[{Ostro {et~al.}(2002)Ostro, Hudson, Benner, Giorgini, Magri, Margot,
  \& Nolan}]{Ostro2002}
Ostro, S.~J., Hudson, R.~S., Benner, L. A.~M., {et~al.} 2002, in Asteroids III,
  ed. W.~F. Bottke, Jr., A.~Cellino, P.~Paolicchi, \& R.~P. Binzel (Tucson:
  University of Arizona Press), 151--168

\bibitem[{{Palmer} {et~al.}(2022){Palmer}, {Gaskell}, {Daly}, {Barnouin},
  {Adam}, \& {Lauretta}}]{Palmer2022}
{Palmer}, E.~E., {Gaskell}, R., {Daly}, M.~G., {et~al.} 2022, PSJ, 3, 102

\bibitem[{Park {et~al.}(2019)Park, Vaughan, Konopliv, Ermakov, Mastrodemos,
  Castillo-Rogez, Joy, Nathues, Polanskey, Rayman, Riedel, Raymond, Russell, \&
  Zuber}]{Park2019}
Park, R.~S., Vaughan, A.~T., Konopliv, A.~S., {et~al.} 2019, Icarus, 319, 812

\bibitem[{{Preusker} {et~al.}(2015){Preusker}, {Scholten}, {Matz},
  {et~al.}}]{Preusker2015}
{Preusker}, F., {Scholten}, F., {Matz}, K.-D., {et~al.} 2015, A\&A, 583, A33

\bibitem[{{Preusker} {et~al.}(2017){Preusker}, {Scholten}, {Matz},
  {et~al.}}]{Preusker2017}
{Preusker}, F., {Scholten}, F., {Matz}, K.-D., {et~al.} 2017, A\&A, 607, L1

\bibitem[{{Rizk} {et~al.}(2018){Rizk}, {Drouet d'Aubigny}, {Golish},
  {et~al.}}]{Rizk2018}
{Rizk}, B., {Drouet d'Aubigny}, C., {Golish}, D., {et~al.} 2018, Space Sci.
  Rev., 214, 26

\bibitem[{{Saito} {et~al.}(2006){Saito}, {Miyamoto}, {Nakamura},
  {et~al.}}]{Saito2006}
{Saito}, J., {Miyamoto}, H., {Nakamura}, R., {et~al.} 2006, Science, 312, 1341

\bibitem[{{Savant Aira} {et~al.}(2025){Savant Aira}, {Facciolo}, \&
  {Ehret}}]{SavantAira2025}
{Savant Aira}, L., {Facciolo}, G., \& {Ehret}, T. 2025, in Proc. IEEE/CVF
  Conference on Computer Vision and Pattern Recognition (CVPR), 5959--5969

\bibitem[{{Sch{\"o}nberger} \& {Frahm}(2016)}]{Schonberger2016}
{Sch{\"o}nberger}, J.~L. \& {Frahm}, J.-M. 2016, in Proc. IEEE Conference on
  Computer Vision and Pattern Recognition (CVPR), 4104

\bibitem[{Sch{\"o}nberger {et~al.}(2016)Sch{\"o}nberger, Zheng, Frahm, \&
  Pollefeys}]{Schonberger2016MVS}
Sch{\"o}nberger, J.~L., Zheng, E., Frahm, J.-M., \& Pollefeys, M. 2016, in
  Proc. European Conference on Computer Vision (ECCV), 501--518

\bibitem[{{Seabrook} {et~al.}(2022){Seabrook}, {Daly}, {Barnouin},
  {et~al.}}]{Seabrook2022}
{Seabrook}, J.~A., {Daly}, M.~G., {Barnouin}, O.~S., {et~al.} 2022, PSJ, 3, 265

\bibitem[{Sugita {et~al.}(2019)Sugita, Honda, Morota, Kameda, Sawada, Tatsumi,
  Yamada, Honda, Yokota, Kouyama, Sakatani, Ogawa, Suzuki, Okada, Namiki,
  Tanaka, Iijima, Yoshioka, Hayakawa, Cho, Matsuoka, Hirata, Hirata, Miyamoto,
  Domingue, Hirabayashi, Nakamura, Hiroi, Michikami, Michel, Ballouz, Barnouin,
  Ernst, Schr{\"o}der, Kikuchi, Hemmi, Komatsu, Fukuhara, Taguchi, Arai,
  Senshu, Demura, Ogawa, Shimaki, Sekiguchi, M{\"u}ller, Hagermann, Mizuno,
  Noda, Matsumoto, Yamada, Ishihara, Ikeda, Araki, Yamamoto, Abe, Yoshida,
  Higuchi, Sasaki, Oshigami, Tsuruta, Asari, Tazawa, Shizugami, Kimura, Otsubo,
  Yabuta, Hasegawa, Ishiguro, Tachibana, Palmer, Gaskell, Le~Corre, Jaumann,
  Otto, Schmitz, Abell, Barucci, Zolensky, Vilas, Thuillet, Sugimoto, Takaki,
  Suzuki, Kamiyoshihara, Okada, Nagata, Fujimoto, Yoshikawa, Yamamoto, Shirai,
  Noguchi, Ogawa, Terui, Kikuchi, Yamaguchi, Oki, Takao, Takeuchi, Ono, Mimasu,
  Yoshikawa, Takahashi, Takei, Fujii, Hirose, Nakazawa, Hosoda, Mori, Shimada,
  Soldini, Iwata, Abe, Yano, Tsukizaki, Ozaki, Nishiyama, Saiki, Watanabe, \&
  Tsuda}]{Sugita2019}
Sugita, S., Honda, R., Morota, T., {et~al.} 2019, Science, 364, eaaw0422

\bibitem[{Tatsumi {et~al.}(2020)Tatsumi, Domingue, Schr{\"o}der, Yokota,
  Kuroda, Ishiguro, Hasegawa, Hiroi, Honda, Hemmi, Le~Corre, Sakatani, Morota,
  Yamada, Kameda, Koyama, Suzuki, Cho, Yoshioka, Matsuoka, Honda, Hayakawa,
  Hirata, Hirata, Yamamoto, Vilas, Takato, Yoshikawa, Abe, \&
  Sugita}]{Tatsumi2020}
Tatsumi, E., Domingue, D., Schr{\"o}der, S., {et~al.} 2020, A\&A, 639, A83

\bibitem[{{Taubin}(1995)}]{Taubin1995}
{Taubin}, G. 1995, in Proc. SIGGRAPH 95, 351

\bibitem[{{Wang} {et~al.}(2021){Wang}, {Liu}, {Liu}, {Theobalt}, {Komura}, \&
  {Wang}}]{Wang2021NeuS}
{Wang}, P., {Liu}, L., {Liu}, Y., {et~al.} 2021, in Advances in Neural
  Information Processing Systems, Vol.~34, 27171--27183

\bibitem[{{Wang} {et~al.}(2004){Wang}, {Bovik}, {Sheikh}, \&
  {Simoncelli}}]{Wang2004SSIM}
{Wang}, Z., {Bovik}, A.~C., {Sheikh}, H.~R., \& {Simoncelli}, E.~P. 2004, IEEE
  Trans. Image Process., 13, 600

\bibitem[{{Watanabe} {et~al.}(2019){Watanabe}, {Hirabayashi}, {Hirata},
  {et~al.}}]{Watanabe2019}
{Watanabe}, S., {Hirabayashi}, M., {Hirata}, N., {et~al.} 2019, Science, 364,
  268

\bibitem[{{Williams}(1978)}]{Williams1978}
{Williams}, L. 1978, in Proc. SIGGRAPH 78, 270

\bibitem[{{Yu} {et~al.}(2024{\natexlab{a}}){Yu}, {Chen}, {Huang}, {Sattler}, \&
  {Geiger}}]{Yu2024Mip}
{Yu}, Z., {Chen}, A., {Huang}, B., {Sattler}, T., \& {Geiger}, A.
  2024{\natexlab{a}}, in Proc. IEEE/CVF Conference on Computer Vision and
  Pattern Recognition (CVPR), 19447--19456

\bibitem[{{Yu} {et~al.}(2024{\natexlab{b}}){Yu}, {Sattler}, \&
  {Geiger}}]{Yu2024GOF}
{Yu}, Z., {Sattler}, T., \& {Geiger}, A. 2024{\natexlab{b}}, ACM Trans. Graph.,
  43, 271

\bibitem[{{Zhang} {et~al.}(2024){Zhang}, {Fang}, {Shrestha},
  {et~al.}}]{Zhang2024RaDe}
{Zhang}, B., {Fang}, C., {Shrestha}, R., {et~al.} 2024, arXiv e-prints,
  arXiv:2406.01467

\bibitem[{{Zhang} {et~al.}(2018){Zhang}, {Isola}, {Efros}, {Shechtman}, \&
  {Wang}}]{Zhang2018LPIPS}
{Zhang}, R., {Isola}, P., {Efros}, A.~A., {Shechtman}, E., \& {Wang}, O. 2018,
  in Proc. IEEE/CVF Conference on Computer Vision and Pattern Recognition
  (CVPR), 586--595

\bibitem[{{Zhang} {et~al.}(2025){Zhang}, {Cui}, {Wei}, \&
  {Chen}}]{Zhang2025AsteroidGS}
{Zhang}, X., {Cui}, L., {Wei}, X., \& {Chen}, Y. 2025, A\&A, 704, A247

\bibitem[{Zhang {et~al.}(2021)Zhang, Srinivasan, Deng, Debevec, Freeman, \&
  Barron}]{Zhang2021NeRFactor}
Zhang, X., Srinivasan, P.~P., Deng, B., {et~al.} 2021, ACM Trans. Graph., 40,
  237

\end{thebibliography}

\begin{appendix}
\renewcommand{\topfraction}{0.98}
\renewcommand{\floatpagefraction}{0.98}
\setlength{\floatsep}{8pt plus 2pt minus 2pt}
\twocolumn[{%
\section{Extended benchmark and ablation tables}
\label{app:tables}
\noindent This appendix provides the extended shape and novel-view synthesis metrics on Itokawa and Ryugu
(Tables~\ref{tab:shape_full} and~\ref{tab:images_full}), together with two single-factor diagnostics stepping through
the pre-smoothing mesh-extraction stages (Table~\ref{tab:pipeline}) and isolating the phase switch without terminator
masking (Table~\ref{tab:priors}).
\par\vspace{4pt}%
{\renewcommand{\arraystretch}{0.89}\noindent\begin{minipage}{\textwidth}
\settablecaptype
\centering
\caption{Extended shape evaluation metrics on Itokawa and Ryugu (mean $\pm$ standard deviation over three splits, evaluated under the protocol of Table~\ref{tab:main}).}
\label{tab:shape_full}
\scriptsize
\setlength{\tabcolsep}{4.46pt}
\begin{tabular}{lcccccccccc}
\toprule
Method & Mean (m) & RMSE (m) & Std (m) & Chamfer (m) & Acc. & Compl. & F@0.5~m & F@1~m & F@2~m & Normal ($^\circ$) \\
\midrule
\multicolumn{11}{l}{\textit{Itokawa (AMICA, phase angle 10$^\circ$)}} \\
COLMAP MVS & 0.85\,$\pm$\,0.01 & 1.49\,$\pm$\,0.04 & 1.23\,$\pm$\,0.05 & 0.71\,$\pm$\,0.01 & 0.791\,$\pm$\,0.003 & \underline{0.825}\,$\pm$\,0.003 & 0.527\,$\pm$\,0.003 & 0.808\,$\pm$\,0.003 & 0.950\,$\pm$\,0.002 & 15.1\,$\pm$\,0.1 \\
2DGS & 6.55\,$\pm$\,3.41 & 17.96\,$\pm$\,11.62 & 16.70\,$\pm$\,11.16 & 4.63\,$\pm$\,1.92 & 0.503\,$\pm$\,0.047 & 0.598\,$\pm$\,0.018 & 0.338\,$\pm$\,0.024 & 0.546\,$\pm$\,0.035 & 0.711\,$\pm$\,0.037 & 24.1\,$\pm$\,2.4 \\
GOF & 2.40\,$\pm$\,0.12 & 4.66\,$\pm$\,0.45 & 4.00\,$\pm$\,0.46 & 3.01\,$\pm$\,0.24 & 0.282\,$\pm$\,0.002 & 0.502\,$\pm$\,0.010 & 0.196\,$\pm$\,0.004 & 0.362\,$\pm$\,0.004 & 0.593\,$\pm$\,0.006 & 41.5\,$\pm$\,0.4 \\
RaDe-GS & \underline{0.62}\,$\pm$\,0.01 & \textbf{0.94}\,$\pm$\,0.03 & \textbf{0.71}\,$\pm$\,0.03 & \underline{0.66}\,$\pm$\,0.00 & \underline{0.840}\,$\pm$\,0.004 & 0.817\,$\pm$\,0.002 & \underline{0.542}\,$\pm$\,0.004 & \underline{0.828}\,$\pm$\,0.003 & \underline{0.962}\,$\pm$\,0.001 & \underline{13.6}\,$\pm$\,0.2 \\
PGSR & 0.99\,$\pm$\,0.07 & 2.83\,$\pm$\,0.46 & 2.66\,$\pm$\,0.46 & 0.76\,$\pm$\,0.03 & 0.807\,$\pm$\,0.002 & 0.818\,$\pm$\,0.002 & 0.526\,$\pm$\,0.001 & 0.812\,$\pm$\,0.001 & 0.958\,$\pm$\,0.001 & 14.3\,$\pm$\,0.1 \\
Asteroid-GS & 1.01\,$\pm$\,0.10 & 2.10\,$\pm$\,0.41 & 1.83\,$\pm$\,0.41 & 1.76\,$\pm$\,0.55 & 0.635\,$\pm$\,0.040 & 0.665\,$\pm$\,0.021 & 0.376\,$\pm$\,0.022 & 0.649\,$\pm$\,0.031 & 0.835\,$\pm$\,0.028 & 13.9\,$\pm$\,1.1 \\
Asteroid-GS$^{\dagger}$ & 0.94\,$\pm$\,0.03 & 1.82\,$\pm$\,0.17 & 1.55\,$\pm$\,0.19 & 1.54\,$\pm$\,0.16 & 0.659\,$\pm$\,0.008 & 0.658\,$\pm$\,0.008 & 0.378\,$\pm$\,0.003 & 0.658\,$\pm$\,0.008 & 0.848\,$\pm$\,0.007 & \textbf{12.7}\,$\pm$\,0.2 \\
Mip-Splatting$^{\S}$ & 2.48 & 3.20 & 2.01 & 73.26 & 0.315 & 0.105 & 0.086 & 0.158 & 0.247 & 50.0 \\
\rowcolor{pcgsrow} \textbf{PCGS (this work)} & \textbf{0.59}\,$\pm$\,0.02 & \underline{1.02}\,$\pm$\,0.15 & \underline{0.83}\,$\pm$\,0.17 & \textbf{0.59}\,$\pm$\,0.01 & \textbf{0.847}\,$\pm$\,0.007 & \textbf{0.859}\,$\pm$\,0.004 & \textbf{0.579}\,$\pm$\,0.005 & \textbf{0.853}\,$\pm$\,0.005 & \textbf{0.969}\,$\pm$\,0.002 & 14.7\,$\pm$\,0.2 \\
\addlinespace[1.5pt]
\multicolumn{11}{l}{\textit{Ryugu (ONC-T, phase angle 19$^\circ$)}} \\
COLMAP MVS & 1.13\,$\pm$\,0.01 & 1.76\,$\pm$\,0.04 & 1.35\,$\pm$\,0.04 & 0.87\,$\pm$\,0.01 & 0.719\,$\pm$\,0.004 & 0.717\,$\pm$\,0.003 & 0.436\,$\pm$\,0.004 & 0.718\,$\pm$\,0.003 & 0.916\,$\pm$\,0.002 & 11.9\,$\pm$\,0.1 \\
2DGS & 2.50\,$\pm$\,0.08 & 3.80\,$\pm$\,0.17 & 2.86\,$\pm$\,0.16 & 1.96\,$\pm$\,0.07 & 0.373\,$\pm$\,0.011 & 0.471\,$\pm$\,0.008 & 0.222\,$\pm$\,0.006 & 0.416\,$\pm$\,0.010 & 0.679\,$\pm$\,0.008 & 28.0\,$\pm$\,0.4 \\
GOF & 4.20\,$\pm$\,0.08 & 5.98\,$\pm$\,0.15 & 4.25\,$\pm$\,0.14 & 4.50\,$\pm$\,0.11 & 0.129\,$\pm$\,0.003 & 0.220\,$\pm$\,0.003 & 0.082\,$\pm$\,0.002 & 0.162\,$\pm$\,0.003 & 0.319\,$\pm$\,0.006 & 42.4\,$\pm$\,0.3 \\
RaDe-GS & 3.12\,$\pm$\,0.85 & 4.19\,$\pm$\,1.23 & 2.79\,$\pm$\,0.90 & 3.43\,$\pm$\,0.86 & 0.223\,$\pm$\,0.063 & 0.286\,$\pm$\,0.028 & 0.128\,$\pm$\,0.025 & 0.249\,$\pm$\,0.050 & 0.456\,$\pm$\,0.085 & 32.6\,$\pm$\,7.7 \\
PGSR & \underline{0.73}\,$\pm$\,0.01 & \underline{1.18}\,$\pm$\,0.04 & \underline{0.93}\,$\pm$\,0.04 & \underline{0.76}\,$\pm$\,0.01 & \underline{0.791}\,$\pm$\,0.005 & \underline{0.782}\,$\pm$\,0.005 & \underline{0.499}\,$\pm$\,0.007 & \underline{0.787}\,$\pm$\,0.005 & \textbf{0.949}\,$\pm$\,0.001 & 14.5\,$\pm$\,0.1 \\
Asteroid-GS & 1.27\,$\pm$\,0.02 & 1.88\,$\pm$\,0.03 & 1.39\,$\pm$\,0.03 & 1.42\,$\pm$\,0.03 & 0.532\,$\pm$\,0.006 & 0.518\,$\pm$\,0.005 & 0.284\,$\pm$\,0.004 & 0.525\,$\pm$\,0.005 & 0.797\,$\pm$\,0.005 & \underline{11.4}\,$\pm$\,0.0 \\
Asteroid-GS$^{\dagger}$ & 1.29\,$\pm$\,0.01 & 1.98\,$\pm$\,0.02 & 1.51\,$\pm$\,0.02 & 1.52\,$\pm$\,0.02 & 0.537\,$\pm$\,0.004 & 0.523\,$\pm$\,0.004 & 0.284\,$\pm$\,0.002 & 0.530\,$\pm$\,0.004 & 0.794\,$\pm$\,0.004 & \textbf{11.3}\,$\pm$\,0.0 \\
\rowcolor{pcgsrow} \textbf{PCGS (this work)} & \textbf{0.62}\,$\pm$\,0.00 & \textbf{0.89}\,$\pm$\,0.01 & \textbf{0.65}\,$\pm$\,0.01 & \textbf{0.73}\,$\pm$\,0.00 & \textbf{0.802}\,$\pm$\,0.000 & \textbf{0.794}\,$\pm$\,0.000 & \textbf{0.516}\,$\pm$\,0.001 & \textbf{0.798}\,$\pm$\,0.000 & \underline{0.948}\,$\pm$\,0.001 & 12.1\,$\pm$\,0.0 \\
\bottomrule
\end{tabular}
\tablefoot{Std is the standard deviation of unsigned vertex distances; Acc.\ and Compl.\ are accuracy and completeness at 1~m (Eq.~\eqref{eq:pr}). $^{\dagger}$Default camera-ring scene extent. $^{\S}$First split only (TSDF depth fusion).}
\end{minipage}
}%
\par\vspace{6pt}%
}]

\begin{table}[!htbp]
\centering
\caption{Novel-view synthesis metrics on the test images of Itokawa and Ryugu (mean $\pm$ standard deviation over three splits).}
\label{tab:images_full}
\scriptsize
\setlength{\tabcolsep}{6.60pt}
\begin{tabular}{lcccc}
\toprule
Method & PSNR$_\mathrm{raw}$ & PSNR (dB) & SSIM & LPIPS \\
\midrule
\multicolumn{5}{l}{\textit{Itokawa (AMICA, phase angle 10$^\circ$)}} \\
2DGS & 33.1\,$\pm$\,1.1 & 33.6\,$\pm$\,1.0 & 0.968\,$\pm$\,0.005 & 0.109\,$\pm$\,0.011 \\
GOF & 38.0\,$\pm$\,0.6 & 38.0\,$\pm$\,0.6 & 0.973\,$\pm$\,0.002 & 0.109\,$\pm$\,0.011 \\
RaDe-GS & \underline{39.9}\,$\pm$\,0.9 & \underline{40.0}\,$\pm$\,0.9 & \underline{0.983}\,$\pm$\,0.001 & \textbf{0.096}\,$\pm$\,0.008 \\
PGSR & 39.2\,$\pm$\,0.8 & 39.3\,$\pm$\,0.8 & 0.982\,$\pm$\,0.001 & \textbf{0.096}\,$\pm$\,0.008 \\
Asteroid-GS & 16.7\,$\pm$\,0.8 & 29.6\,$\pm$\,0.9 & 0.926\,$\pm$\,0.010 & 0.144\,$\pm$\,0.009 \\
Asteroid-GS$^{\dagger}$ & 17.0\,$\pm$\,0.9 & 29.5\,$\pm$\,0.5 & 0.943\,$\pm$\,0.004 & 0.138\,$\pm$\,0.011 \\
Mip-Splatting$^{\S}$ & \textbf{42.5} & \textbf{42.6} & \textbf{0.985} & 0.100 \\
\rowcolor{pcgsrow} \textbf{PCGS (this work)} & 36.4\,$\pm$\,0.6 & 36.5\,$\pm$\,0.5 & 0.980\,$\pm$\,0.001 & \underline{0.098}\,$\pm$\,0.008 \\
\addlinespace[1.5pt]
\multicolumn{5}{l}{\textit{Ryugu (ONC-T, phase angle 19$^\circ$)}} \\
2DGS & 34.8\,$\pm$\,1.8 & 34.9\,$\pm$\,1.8 & 0.948\,$\pm$\,0.005 & \underline{0.225}\,$\pm$\,0.001 \\
GOF & 33.5\,$\pm$\,0.0 & 33.5\,$\pm$\,0.0 & 0.944\,$\pm$\,0.001 & \textbf{0.202}\,$\pm$\,0.008 \\
RaDe-GS & \textbf{35.4}\,$\pm$\,1.4 & \textbf{35.8}\,$\pm$\,0.9 & \underline{0.950}\,$\pm$\,0.004 & 0.231\,$\pm$\,0.005 \\
PGSR & \underline{35.1}\,$\pm$\,0.2 & \underline{35.2}\,$\pm$\,0.2 & 0.947\,$\pm$\,0.003 & 0.227\,$\pm$\,0.002 \\
Asteroid-GS & 13.4\,$\pm$\,0.6 & 22.7\,$\pm$\,0.2 & 0.858\,$\pm$\,0.002 & 0.283\,$\pm$\,0.002 \\
Asteroid-GS$^{\dagger}$ & 13.5\,$\pm$\,0.6 & 22.7\,$\pm$\,0.0 & 0.868\,$\pm$\,0.004 & 0.282\,$\pm$\,0.002 \\
\rowcolor{pcgsrow} \textbf{PCGS (this work)} & 31.0\,$\pm$\,0.1 & 31.3\,$\pm$\,0.1 & \textbf{0.952}\,$\pm$\,0.001 & \underline{0.225}\,$\pm$\,0.002 \\
\bottomrule
\end{tabular}
\tablefoot{PSNR$_\mathrm{raw}$ (dB) is evaluated on raw renders; the remaining columns apply per-image linear gain-and-offset matching over the foreground mask. COLMAP MVS does not synthesise novel views and is omitted. $^{\dagger}$Default camera-ring scene extent; $^{\S}$first split only.}
\end{table}

\begin{table}[!htbp]
\caption{Step-by-step ablation of the mesh extraction pipeline and pixel-convention alignment.}
\label{tab:pipeline}
\centering
\scriptsize\renewcommand{\arraystretch}{0.90}\setlength{\tabcolsep}{0pt}
\begin{tabular*}{\columnwidth}{@{\extracolsep{\fill}}lcccccc@{}}
\toprule
 & \multicolumn{3}{c}{Itokawa} & \multicolumn{3}{c}{Ryugu, 19$^\circ$} \\
\cmidrule(lr){2-4}\cmidrule(lr){5-7}
Pipeline stage & Mean & RMSE & $<$1~m & Mean & RMSE & $<$1~m \\
 & (m) & (m) & & (m) & (m) & \\
\midrule
Initial Poisson extraction & 0.78 & 1.56 & 0.769 & 1.35 & 1.66 & 0.408 \\
+ pixel conv.\ (extraction) & 0.68 & 1.52 & 0.833 & 0.94 & 1.24 & 0.629 \\
+ pixel conv.\ (training)   & 0.69 & 1.71 & 0.843 & 0.76 & 1.08 & 0.735 \\
+ biharmonic fairing        & 0.65 & 1.29 & 0.843 & 0.76 & 1.06 & 0.734 \\
+ tie-point normal offset   & 0.64 & 1.29 & 0.848 & 0.66 & 0.95 & 0.803 \\
\bottomrule
\end{tabular*}
\tablefoot{Shading-only three-split means before evaluation smoothing; $<$1~m is the fraction of reconstructed vertices
within 1~m of the reference.}
\end{table}

\begin{table}[!htbp]
\centering
\caption{Effect of the total-variation albedo prior and cast-shadow mapping on the first split.}
\label{tab:priors}
\scriptsize
\renewcommand{\arraystretch}{1.04}
\setlength{\tabcolsep}{4.04pt}
\begin{tabular}{lcrrrrr}
\toprule
Variant & Switch & Mean & RMSE & F@1~m & Normal & PSNR \\
 & & (m) & (m) & & ($^\circ$) & (dB) \\
\midrule
\multicolumn{7}{l}{\textit{Itokawa, 10$^\circ$}} \\
Shading only &  & 0.66 & 1.46 & 0.827 & \underline{14.6} & \textbf{38.1} \\
+ albedo prior & \checkmark & \textbf{0.58} & \textbf{0.92} & \underline{0.846} & \textbf{14.4} & 36.8 \\
\rowcolor{pcgsrow} \textbf{PCGS (final model)} &  & \underline{0.60} & \underline{1.07} & \textbf{0.849} & 14.8 & \underline{37.2} \\
\addlinespace[3pt]
\multicolumn{7}{l}{\textit{Ryugu, 7$^\circ$}} \\
Shading only &  & \textbf{0.74} & \textbf{1.04} & \textbf{0.719} & \textbf{12.6} & \textbf{34.8} \\
+ albedo prior & \checkmark & 0.76 & \textbf{1.04} & 0.708 & \underline{12.7} & \underline{33.7} \\
\rowcolor{pcgsrow} \textbf{PCGS (final model)} &  & \underline{0.75} & \underline{1.05} & \underline{0.712} & 13.0 & 33.5 \\
\addlinespace[3pt]
\multicolumn{7}{l}{\textit{Ryugu, 19$^\circ$}} \\
Shading only &  & 0.65 & 0.95 & 0.770 & 12.4 & \textbf{32.5} \\
+ albedo prior &  & 0.66 & 0.96 & 0.767 & 13.1 & 31.0 \\
+ cast shadows & \checkmark & \textbf{0.62} & \underline{0.90} & \underline{0.791} & \textbf{11.9} & 30.0 \\
+ albedo prior + cast shadows &  & \underline{0.63} & 0.93 & 0.785 & 12.4 & 28.6 \\
\rowcolor{pcgsrow} \textbf{PCGS (final model)} &  & \textbf{0.62} & \textbf{0.89} & \textbf{0.798} & \underline{12.1} & \underline{31.4} \\
\addlinespace[3pt]
\multicolumn{7}{l}{\textit{Ryugu, 40$^\circ$}} \\
Shading only &  & \underline{0.71} & \underline{1.06} & \underline{0.719} & 13.2 & \textbf{33.9} \\
+ cast shadows & \checkmark & \underline{0.71} & 1.07 & 0.709 & \textbf{12.5} & 28.1 \\
+ albedo prior + cast shadows &  & 0.77 & 1.24 & 0.676 & 13.0 & 26.6 \\
\rowcolor{pcgsrow} \textbf{PCGS (final model)} &  & \textbf{0.66} & \textbf{0.95} & \textbf{0.757} & \underline{12.8} & \underline{33.0} \\
\addlinespace[3pt]
\multicolumn{7}{l}{\textit{Bennu, 26--60$^\circ$}} \\
Shading only &  & 14.75 & 27.49 & 0.271 & 38.6 & \underline{20.3} \\
+ cast shadows & \checkmark & \underline{4.18} & \underline{10.28} & \underline{0.444} & \underline{34.8} & 19.8 \\
\rowcolor{pcgsrow} \textbf{PCGS (final model)} &  & \textbf{0.51} & \textbf{0.81} & \textbf{0.788} & \textbf{31.9} & \textbf{23.4} \\
\bottomrule
\end{tabular}
\tablefoot{All rows except PCGS (final model) omit the night-side band; \checkmark\ marks the branch selected by the phase switch.}
\end{table}

\begin{figure*}[!p]
\centering
\includegraphics[width=\textwidth]{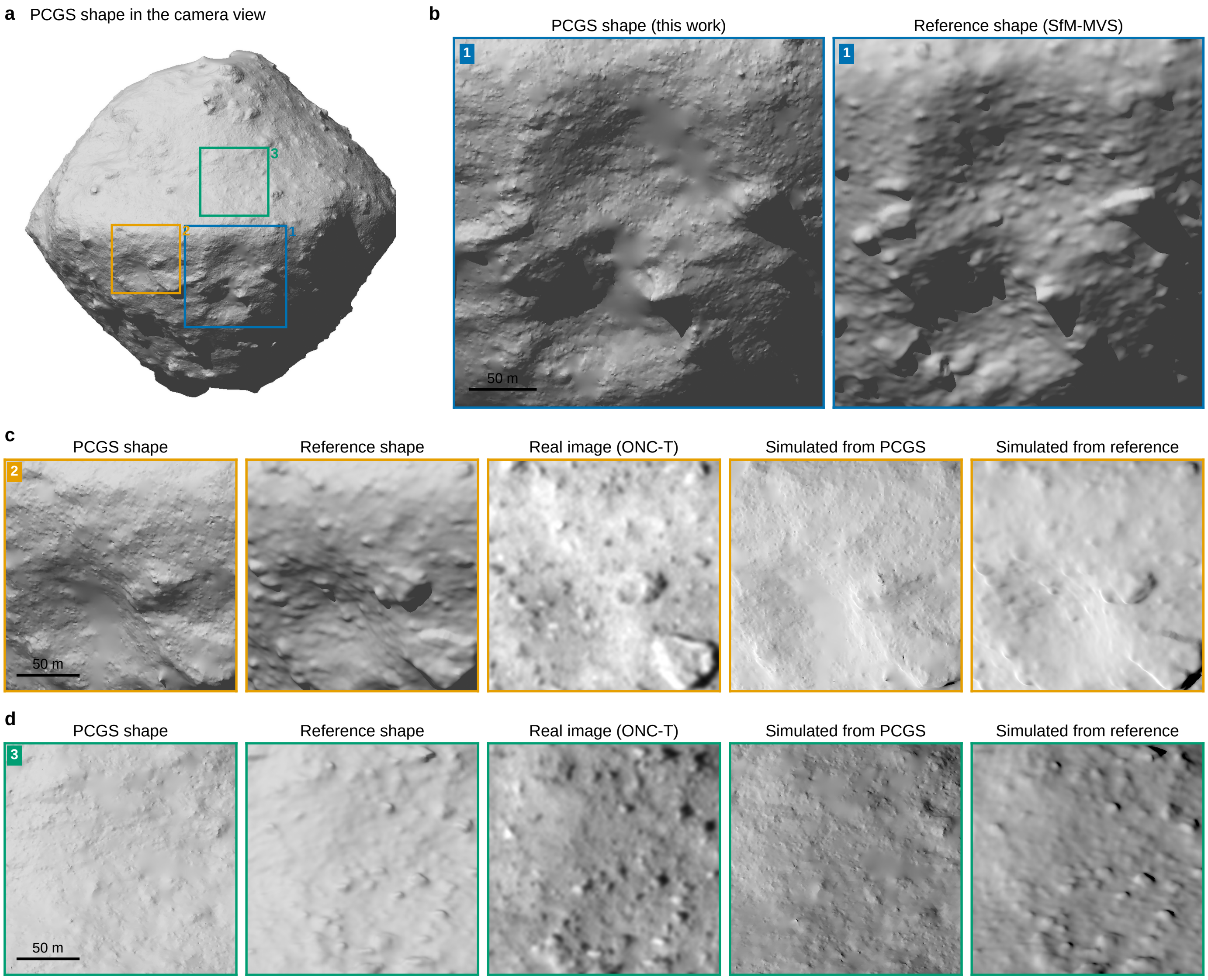}
\caption{Local boulder-scale comparisons on Ryugu at high solar phase angle ($\alpha=39.9^\circ$) for held-out
Hayabusa2 ONC-T test image \texttt{hyb2\_onc\_20180831\_070008\_tvf\_l2c} (2018 August 31, 07:00 UTC; ground pixel scale
$2.29$--$2.30$~m~px$^{-1}$). Regions 1--3 (50~m scale bars) compare the \pcgs\ shape model, the mission SfM-MVS
reference shape, the real ONC-T image, and lunar-Lambert simulated renders produced from both shape models.}
\label{fig:detail_ryugu40}
\end{figure*}

\clearpage
\twocolumn[{%
\section{Local boulder-scale comparisons on Ryugu}
\label{app:details}
\noindent Figures~\ref{fig:detail_ryugu19} and~\ref{fig:detail_ryugu40} provide local boulder- and crater-scale
qualitative comparisons on held-out Hayabusa2 ONC-T test views of Ryugu at moderate ($\alpha=18.5^\circ$) and high
($\alpha=39.9^\circ$) solar phase angles. Across three representative surface regions in each epoch, the \pcgs\ shape
model is compared directly against the mission SfM-MVS reference shape \citep{Watanabe2019}, the real spacecraft image,
and lunar-Lambert synthetic renders generated from both shapes under identical viewing and illumination geometries.
\par\vspace{8pt}%
\noindent\begin{minipage}{\textwidth}%
\setfigcaptype
\centering
\includegraphics[width=\textwidth]{\figfile{figS4_detail_ryugu19}}
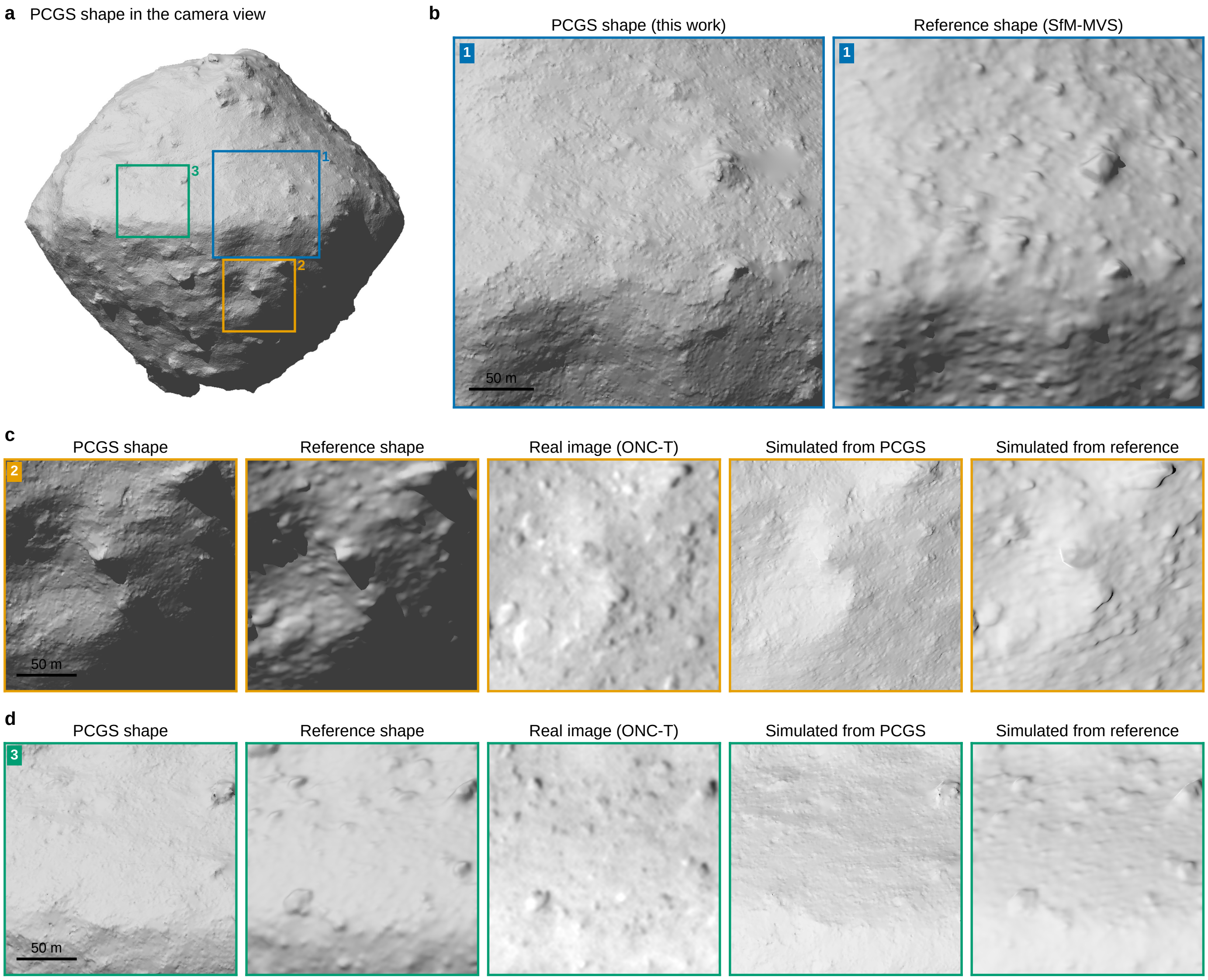
\caption{Local boulder-scale comparisons on Ryugu at moderate solar phase angle ($\alpha=18.5^\circ$) for held-out
Hayabusa2 ONC-T test image \texttt{hyb2\_onc\_20180710\_113340\_tvf\_l2c} (2018 July 10, 11:33 UTC; ground pixel scale
$2.11$--$2.12$~m~px$^{-1}$). Regions 1--3 (50~m scale bars) compare the \pcgs\ shape model, the mission SfM-MVS
reference shape, the real ONC-T image, and lunar-Lambert simulated renders produced from both shape models.}
\label{fig:detail_ryugu19}
\end{minipage}%
\par\vspace{4pt}%
}]
\null\clearpage

\end{appendix}

\end{document}